\documentclass[lettersize,journal]{IEEEtran}

\usepackage{amsmath,amssymb,amsfonts}

\usepackage{algorithmic}
\usepackage[linesnumbered,ruled,vlined]{algorithm2e}
\SetKwInput{KwIn}{Input}     
\SetKwInput{KwOut}{Output}  
\SetKwIF{If}{ElseIf}{Else}{if}{then}{else if}{else}{end if}
\SetKwFor{For}{for}{do}{end for}
\SetKw{KwAnd}{and}

\usepackage{graphicx}
\usepackage[compatibility=false]{caption}
\usepackage{subcaption}

\usepackage{booktabs}
\usepackage{multirow}
\usepackage{array}
\usepackage{siunitx}

\usepackage{textcomp}
\usepackage{stfloats}
\usepackage{url}
\usepackage{verbatim}
\usepackage{cite}
\begin{document}

\title{Coverage Path Planning for Autonomous Sailboats in Inhomogeneous and Time-Varying Oceans: A Spatiotemporal Optimization Approach}

\author{Yang An, Zhikang Ge, Taiyu Zhang, Jean-Baptiste R. G. Souppez, Gaofei Xu, Zhengru Ren$^{*}$
\thanks{Manuscript received X; revised X; accepted X. This work was supported by the Shenzhen Science and Technology Program, China, under Grant WDZC20231128135104001. Paper no. X. (Corresponding author: Zhengru Ren.)}
\thanks{Yang An is with the Institute of Deep-sea Science and Engineering, Chinese Academy of Sciences, Sanya, China, and also with the Institute for Ocean Engineering, Shenzhen International Graduate School, Tsinghua University, Shenzhen, China (e-mail: anyang@idsse.ac.cn).}
\thanks{Taiyu Zhang is with the Institute for Ocean Engineering, Shenzhen International Graduate School, Tsinghua University, Shenzhen, China (e-mail: zhangtai22@mails.tsinghua.edu.cn).}
\thanks{Zhikang Ge is with ZJU-Hangzhou Global Scientific and Technological Innovation Center, Zhejiang University, Hangzhou, China (e-mail: zge@zju.edu.cn).} 
\thanks{Jean-Baptiste R. G. Souppez is with the Department of Engineering, Faculty of Computing, Engineering and the Built Environment, Birmingham City University, Birmingham, UK (e-mail: jb.souppez@bcu.ac.uk).}
\thanks{Gaofei Xu is with the Institute of Deep-sea Science and Engineering, Chinese Academy of Sciences, Sanya, China (e-mail: xugf@idsse.ac.cn).}
\thanks{Zhengru Ren is with the School of Ocean and Civil Engineering, Shanghai Jiao Tong University, Shanghai, China (e-mail: ren.zhengru@sjtu.edu.cn).}}

\markboth{Journal of \LaTeX\ Class Files,~Vol.~14, No.~8, August~2021}%
{Shell \MakeLowercase{\textit{et al.}}: A Sample Article Using IEEEtran.cls for IEEE Journals}


\maketitle

\begin{abstract}
Autonomous sailboats are well suited for long-duration ocean observation due to their wind-driven endurance. However, their performance is highly anisotropic and strongly influenced by inhomogeneous and time-varying wind and current fields, limiting the effectiveness of existing coverage methods such as boustrophedon sweeping. Planning under these environmental and maneuvering constraints remains underexplored. This paper presents a spatiotemporal coverage path planning framework tailored to autonomous sailboats, combining (1) topology-based morphological constraints in the spatial domain to promote compact and continuous coverage, and (2) forecast-aware look-ahead planning in the temporal domain to anticipate environmental evolution and enable foresighted decision-making. Simulations conducted under stochastic inhomogeneous and time-varying ocean environments, including scenarios with partial directional accessibility, demonstrate that the proposed method generates efficient and feasible coverage paths where traditional strategies often fail. To the best of our knowledge, this study provides the first dedicated solution to the coverage path planning problem for autonomous sailboats operating in inhomogeneous and time-varying ocean environments, establishing a foundation for future cooperative multi-sailboat coverage.
\end{abstract}


\section{Introduction}
\IEEEPARstart{D}{ue} to their superior endurance capabilities, autonomous sailboats have emerged as essential platforms within integrated ocean observation systems~\cite{chaiMonitoringOceanBiogeochemistry2020,silva2019rigid}. They have already been deployed in missions such as acoustic oceanographic surveys~\cite{cabrera-gamez_2019_acoustic} and marine meteorological sensing~\cite{zhangHurricaneObservationsUncrewed2023}, and are expected to play an increasing role in applications including bathymetric mapping, maritime search and rescue, and fishery operations. A fundamental operational requirement across these scenarios is the ability to achieve complete and efficient coverage of designated regions of interest.

\begin{figure*}[!t]
\centering
\begin{subfigure}[t]{0.32\linewidth}
    \centering
    \includegraphics[height=0.75\linewidth, keepaspectratio]{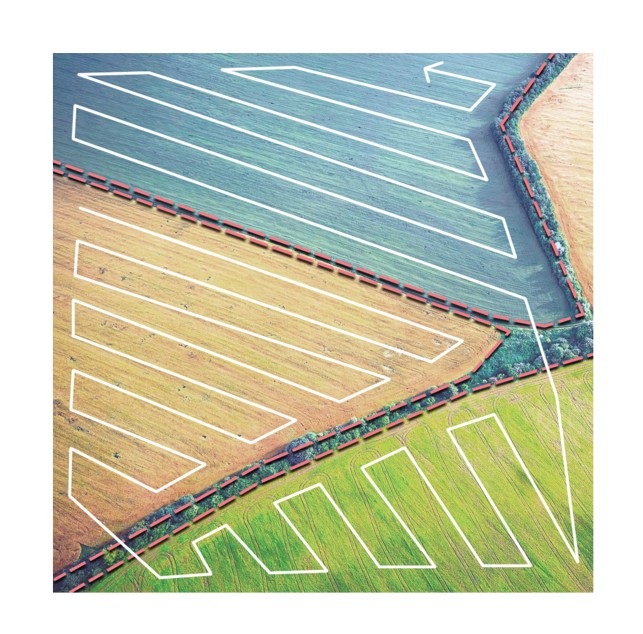}
    \caption{Decomposition and boustrophedon in structured regions}
    \label{FIG:Boustrophedon}
\end{subfigure}
\hfill
\begin{subfigure}[t]{0.32\linewidth}
    \centering
    \includegraphics[height=0.75\linewidth, keepaspectratio]{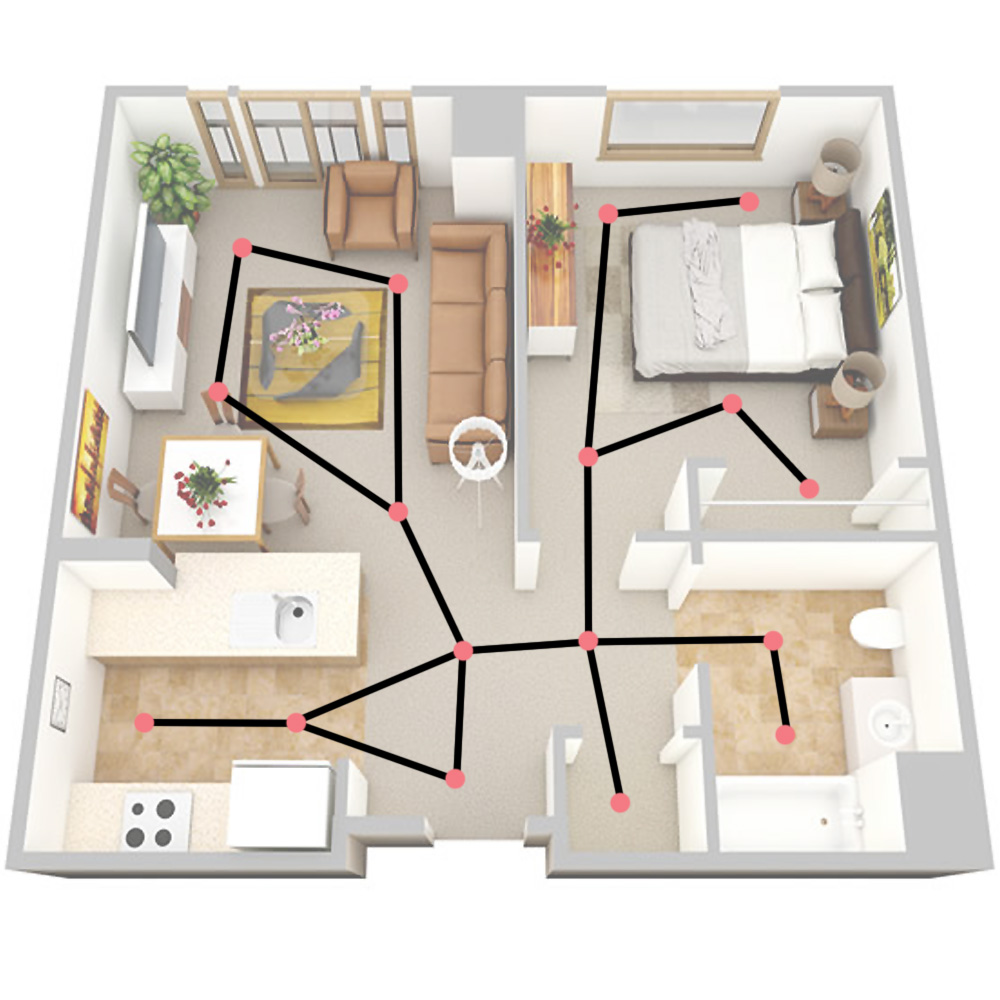}
    \caption{Graph-based method in obstacle-rich environments}
    \label{FIG:Graph-based}
\end{subfigure}
\hfill
\begin{subfigure}[t]{0.32\linewidth}
    \centering
    \includegraphics[height=0.75\linewidth, keepaspectratio]{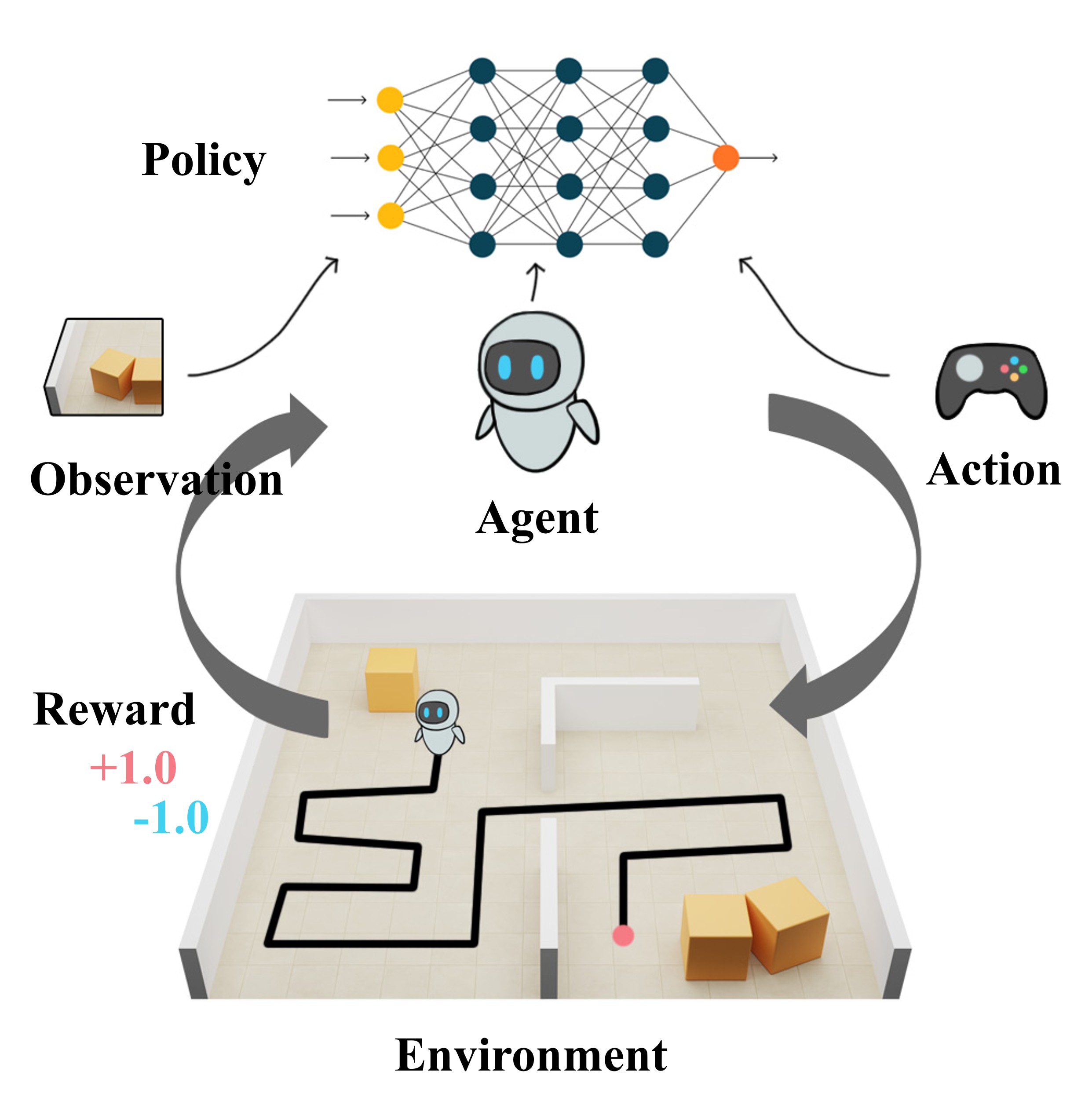}
    \caption{Learning-based method via environmental interaction}
    \label{FIG:RL-based}
\end{subfigure}
\caption{Representative categories of classical CPP strategies.}
\label{FIG:Classic}
\end{figure*}

Coverage path planning (CPP) refers to the problem of generating a continuous trajectory that satisfies one or more of the following criteria~\cite{galceran2013survey}: (1) complete coverage, (2) non-overlapping paths, (3) collision avoidance, (4) simplicity, and (5) optimality under specified constraints. As a classical topic in robotics, CPP has been widely applied in domains such as robotic vacuum cleaning~\cite{hasan2014path}, agricultural robotics~\cite{10050562}, and autonomous underwater vehicles (AUVs) for marine surveys~\cite{galceran2012efficient}. Existing methods can be broadly categorized into three families~\cite{jayalakshmi2025comprehensive,tan2021comprehensive}: region decomposition with systematic sweeping, graph-based representations, and learning-based approaches (Fig.~\ref{FIG:Classic}).

\begin{figure*}[!t]
\centering
\begin{minipage}{0.48\textwidth}
  \centering
  \includegraphics[width=\linewidth,height=0.75\linewidth,keepaspectratio]{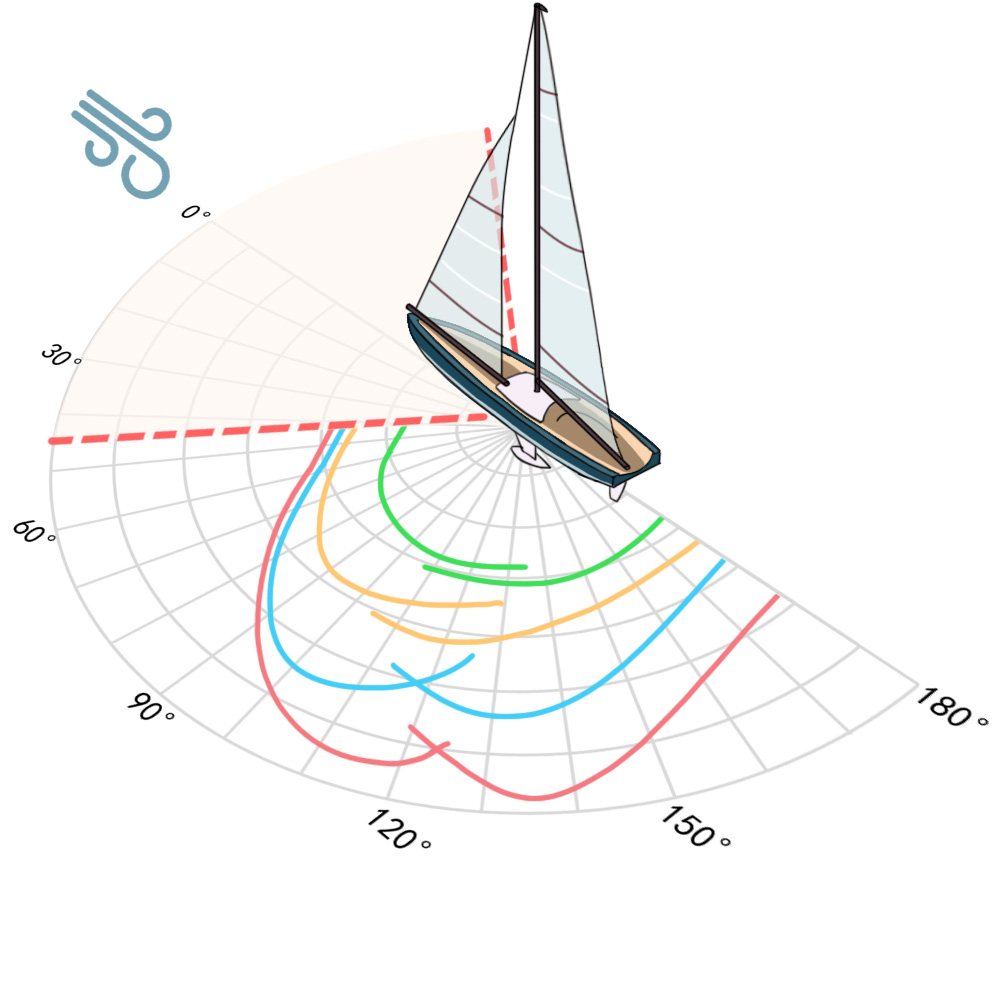}
  \caption*{(a) Direction-dependent sailing efficiency}
\end{minipage}
\hfill
\begin{minipage}{0.48\textwidth}
  \centering
  \includegraphics[width=\linewidth,height=0.75\linewidth,keepaspectratio]{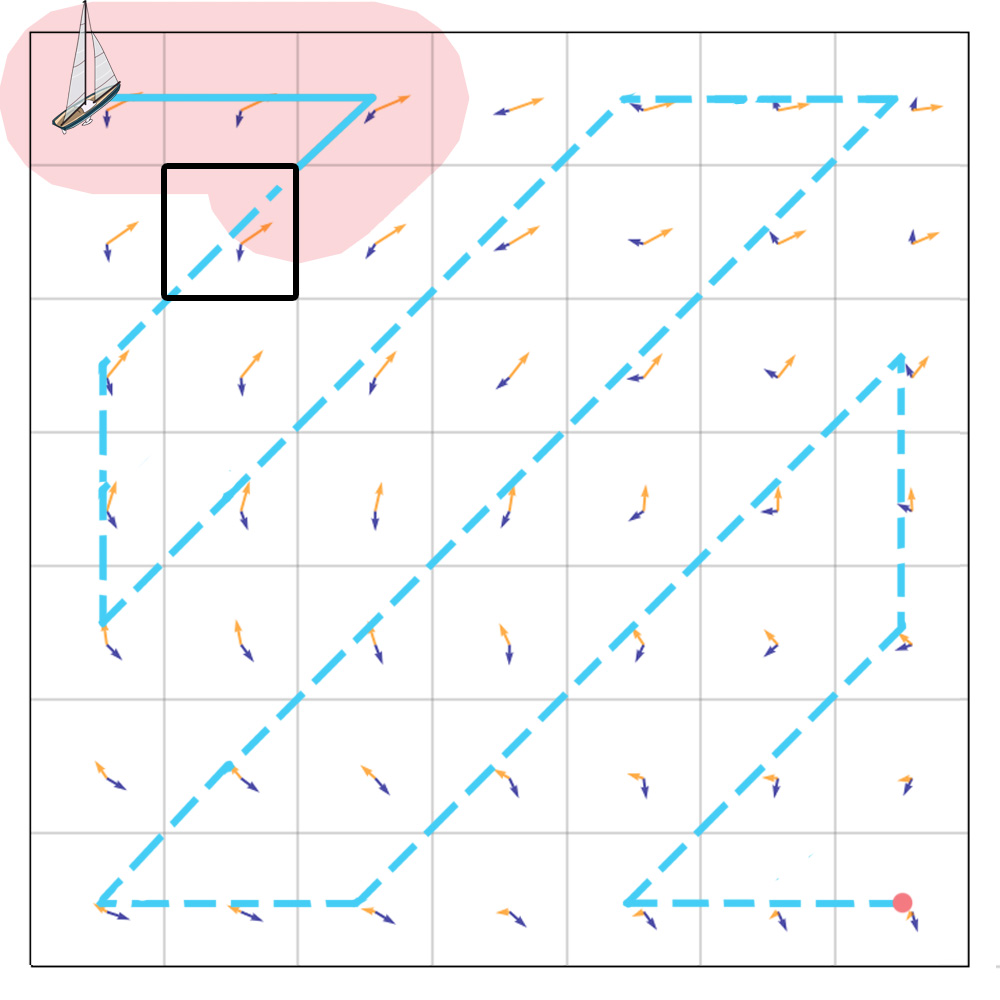}
  \caption*{(b) Temporal partial accessibility}
\end{minipage}
\vspace{0.5em}
\caption{Unique challenges of CPP for autonomous sailboats in time-varying environments. 
(a) The polar plot shows sailboat velocities under varying wind angles and wind speeds. The shaded region represents the no-go zone where direct sailing upwind is infeasible. (b) The black box highlights a region that becomes directionally inaccessible during boustrophedon sweeping due to insufficient effective velocity from wind (blue) and current (orange).}
\label{fig_sim}
\end{figure*}

However, the CPP problem for autonomous sailboats differs fundamentally from traditional robotic platforms due to several unique challenges (Fig.~\ref{fig_sim})~\cite{an2021autonomous}:  
\begin{enumerate}
    \item \textbf{Performance anisotropy:} Unlike typical robots with isotropic motion, autonomous sailboats exhibit strongly anisotropic performance, as sailing efficiency depends on heading relative to wind and current.  
    \item \textbf{Partial accessibility:} Certain regions may only be traversable from specific directions due to currents, eddies, or coastal circulations, making immediate access infeasible and requiring deferred visits when conditions permit.  
    \item \textbf{Environmental variability:} Wind and current fields are both spatially inhomogeneous and temporally varying, defying the static assumptions underlying conventional CPP methods.  
\end{enumerate}

Recent work on sailboat planning has mainly focused on long-horizon point-to-point navigation under dynamic ocean conditions. For example, \cite{sunPathPlanningAlgorithm2024} introduced a deep Q-network–based planner for steady uniform winds, \cite{yuanPathPlanningUnmanned2024} proposed a two-layer A* and potential field method for temporally varying but spatially uniform conditions, and \cite{dengNovelPathPlanning2025} developed a biased-sampling RRT for inhomogeneous and time-varying environments. These efforts demonstrate the feasibility of point-to-point navigation.  

In contrast, coverage-oriented planning remains scarcely studied. The pioneering works of~\cite{shen2025greedy} and \cite{shensailboats2025} addressed single- and multi-sailboat coverage using greedy reward allocation and distributed auction-based coordination, respectively. However, both rely on uniform winds with linear directional changes and overlook flow-induced partial accessibility, limiting applicability in realistic inhomogeneous and time-varying ocean settings.

Beyond sailing, CPP has been extensively studied for unmanned surface vehicles (USVs), unmanned aerial vehicles (UAVs), and mobile robots\cite{maImprovedBABased2022,tranCoveragePathPlanning2023,rekabi-banaUnifiedRobustPath2024}. Representative \textbf{decomposition-and-boustrophedon based strategies} include region division with Traveling Salesman Problem (TSP) ordering and Dubins-curve sweeping for USVs in lakes~\cite{ramkumarsudhaCoveragePathPlanning2024}, and transect orientation optimization for boustrophedon coverage with side-scan sonar~\cite{smithPathCoverageOptimization2022}. In UAV applications, TSP-based ordering combined with boustrophedon coverage and headland paths has been used for spot spraying in irregular fields~\cite{plessenPathPlanningSpot2025}. Yet, such methods rely on static partitioning, fail to predefine subregions due to coverage feasibility depending on both current and future environmental conditions. Direction-dependent accessibility further disrupts regular sweeping or zigzag patterns.

\textbf{Graph-based methods} have also been applied. For example, \cite{chenImprovedCoveragePath2024} integrated BIM-based trapezoidal grid generation with Markov decision processes for indoor robots, while \cite{dongArtificiallyWeightedSpanning2020} proposed an AWSTC algorithm for decentralized multi-robot systems. However, these approaches typically lack the flexibility to encode the need for re-entry from specific directions under changing flow and wind.

Finally, \textbf{reinforcement learning (RL)} methods have gained traction. \cite{wangCoveragePathPlanning2023} applied a re-DQN for orchard harvesting robots, and \cite{xingWideAreaCoveragePath2024} proposed an optimized DQN for deep-sea mining vehicles. While RL offers adaptability to dynamic conditions, it faces exponential state–action growth in inhomogeneous and time-varying oceans, significantly inflating the training burden and limiting practical feasibility in resource-constrained scenarios.

Taken together, these studies demonstrate the breadth of CPP research across domains but also reveal clear limitations when applied to autonomous sailboats, where coverage feasibility is governed by jointly inhomogeneous and time-varying ocean conditions. To the best of our knowledge, no prior work has explicitly formulated and solved the CPP problem under such coupled spatiotemporal constraints. To bridge this gap, we propose a spatiotemporal CPP framework that integrates \emph{morphological constraints} in the spatial domain with \emph{look-ahead planning} in the temporal domain. This dual-constrained approach exploits meteorological forecasts while enforcing spatial compactness, thereby enabling efficient and feasible coverage under realistic marine dynamics.

The main contributions of this work are as follows:  
\begin{enumerate}
    \item We present, to the best of our knowledge, the first CPP framework explicitly formulated for autonomous sailboats in inhomogeneous and time-varying ocean environments, addressing a gap unfilled in prior research.  
    \item We design topology-based morphological constraints that regulate the evolving shape of the covered region, ensuring compactness and continuity while mitigating fragmented coverage and excessive backtracking.  
    \item We incorporate a forecast-aware look-ahead planning strategy that leverages meteorological predictions to anticipate environmental evolution and balance short-term efficiency with long-term feasibility.  
    \item We validate the proposed framework through extensive simulations, demonstrating significant improvements in coverage efficiency and feasibility compared with a baseline boustrophedon strategy under diverse inhomogeneous and time-varying ocean scenarios.  
\end{enumerate}  

The remainder of this paper is organized as follows. Section~II formalizes the environment model, grid structure, action space, and coverage objective. Section~III presents the proposed spatiotemporal optimization framework. Section~IV reports simulation results under inhomogeneous and time-varying ocean environments, highlighting efficiency gains and adaptability. Section~V concludes with a summary and directions for future research.

\section{Problem Formulation}

\subsection{Grid and environment}
A sailboat with anisotropic performance is presented in Fig.\ref{fig_sim}. A few assumptions are summarized as follows: 
\begin{enumerate}

  \item The ocean environment is discretized into a 2D grid map, where each grid cell $g_{ij} \in \mathbb{G}$ is sized such that a sailboat located at the cell center can fully observe the cell with an observation radius $d_{\text{obs}}$. This assumption ensures that one action centered in a cell suffices to mark it as covered. 
  \item The grid size is assumed to be significantly larger than the sailboat's length, allowing full execution of maneuvers such as tacking within a single cell.

  \item Since autonomous sailboats are typically deployed in open-sea scenarios, static and time-varying obstacles are excluded from the map to simplify modeling.

  \item The environment contains time-varying wind and current fields, denoted as $\mathbf{W}_t$ and $\mathbf{C}_t$ respectively. These vector fields define the direction and magnitude of environmental forces at time $t$. The sailboat can only access global environment data through periodically released weather forecasts at an interval of $\Delta t$. Within each cell $g_{ij} \in \mathbb{G}$, wind and current vectors are assumed uniform but may differ between cells. The wind field in cell $g_{ij}$ at time $t$ is denoted $\mathbf{W}_t(i,j)$, and the current field as $\mathbf{C_t}(i,j)$.

  \item It is assumed that the environment remains unchanged within each interval $\Delta t$, and that updates occur only after each action is executed. This reflects the practice that decisions are based on discrete-time observations rather than continuous updates.
  
  \item The forecasts of wind and current fields are subject to cumulative uncertainties. 
  Each predicted field $(\mathbf{W}_{t+k\Delta t}, \mathbf{C}_{t+k\Delta t}), \, k>0$ 
  is regarded as an uncertain estimate rather than a deterministic value. 
  At each forecast step, the predicted flow speed and direction may deviate from the 
  true values by up to $\epsilon_v$ and $\epsilon_\theta$, respectively, and the maximum 
  deviation grows linearly with the prediction horizon, reaching $(k\epsilon_v, k\epsilon_\theta)$ 
  at step $k$. This assumption reflects the cumulative growth of forecast errors in 
  real-world marine environments.

\end{enumerate}

\subsection{Action Definition}

\begin{figure}[!t]
\centering
\scalebox{0.75}{
\includegraphics[width=\linewidth]{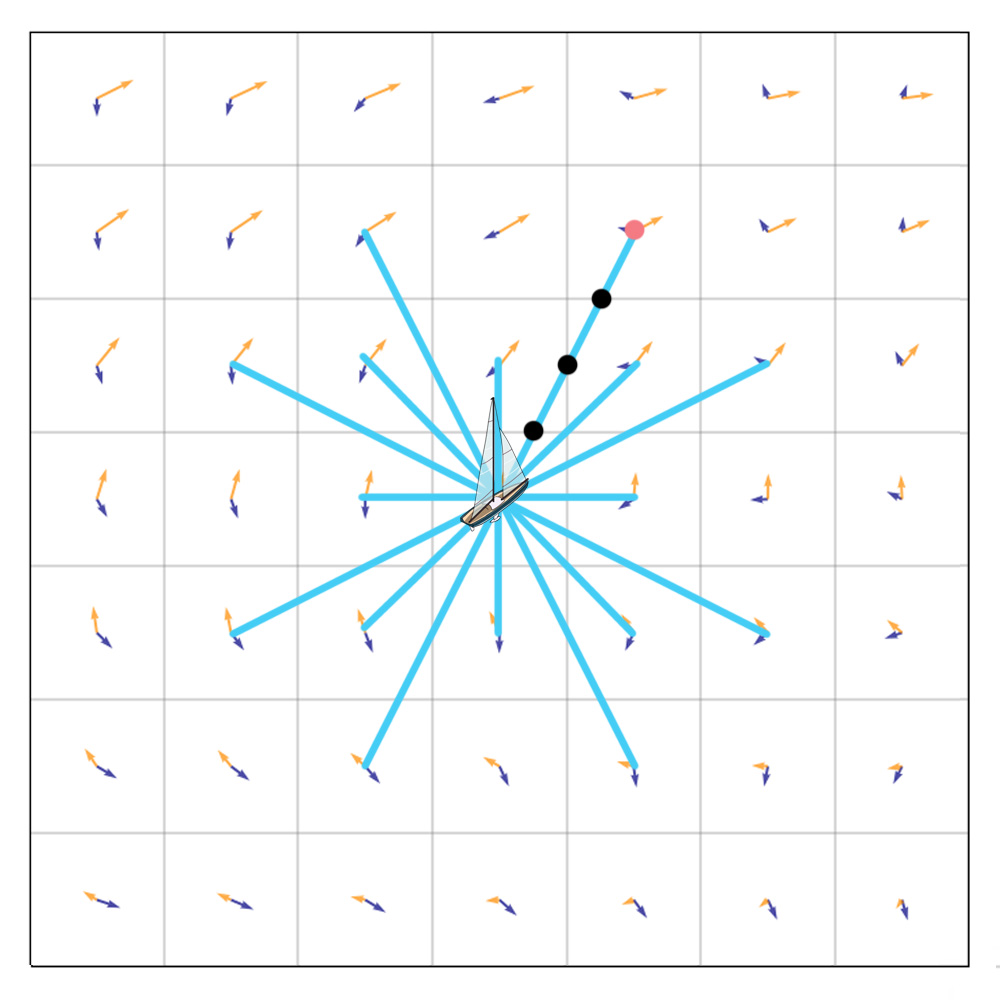}
}
\caption{Illustration of action definitions in autonomous sailboat CPP. The segmented markers indicate that the physical properties of each path segment are computed based on the environmental field of the corresponding grid.}
\label{FIG:Action}
\end{figure}

To capture the anisotropic sailing performance of autonomous sailboats, we define a 16-directional discrete action set \( A = \{a_1, a_2, \ldots, a_{16}\} \), as shown in Fig.~\ref{FIG:Action}, where each action represents a precomputed feasible direction under wind and current constraints. This design avoids the oversimplification of traditional 4- or 8-directional schemes and enables more accurate modeling of the sailboat’s maneuverability. Each action transitions the sailboat from the center of one grid cell to the center of another, ensuring that all waypoints align with grid midpoints. The physical properties encountered along the path are determined by the attributes of the grid cells that the trajectory passes through.

The actual sailing velocity is computed using a velocity prediction program (VPP) calibrated for the 470-class sailboat \cite{masuyama2020science}, as shown in Eq.~\eqref{eq:v_act}. This model offers a realistic yet sufficiently general representation of sailboat performance. Let $\alpha_r$ denote the true wind angle relative to the sailboat’s heading, i.e., the angular difference between the true wind direction and the sailing direction. To reflect the frequent tacking maneuvers required in the no-go zone (typically $\alpha_r < 40^\circ$), the velocity is penalized in this region. To avoid unnecessary complexity while retaining generality, a conservative penalty factor of 0.95 is empirically applied when $\alpha_r \leq 40^\circ$.

\begin{equation}
  \mathbf{V}_{\text{act}} =
  \begin{cases} 
  \mathit{VPP}\Bigl( \mathbf{W}_{t}(i,j),\ \alpha_r \Bigr), & \text{if } \alpha_r > 40^\circ \\
  0.95 \cdot \mathit{VPP}\Bigl( \mathbf{W}_{t}(i,j),\ 40^\circ \Bigr), & \text{if } \alpha_r \leq 40^\circ
  \end{cases}
  \label{eq:v_act}
\end{equation}

 The effective velocity, accounting for environmental flow, is defined as in

  \begin{equation}
  \mathbf{V}_{\text{eff}} = \mathbf{V}_{\text{act}} + \mathbf{C}_{t}(i,j).
  \label{eq:v_eff}
  \end{equation}

An action is considered feasible unless it involves crossing the map boundaries or its execution time exceeds the current phase duration \( \Delta t \).

\subsection{Coverage Objective}

The goal is to minimize the total mission time required to achieve near-complete coverage of the target area. Let \( \gamma(t) \) denote the action trajectory up to time \( t \). The cumulative area covered by this trajectory is defined as:
\begin{equation}
\mathbb{A}(\gamma(t)) = \sum_{g_{ij} \in \mathbb{G}} \mathbb{I}(g_{ij}, \gamma(t)) \, d_{\text{obs}},
\label{eq:coverage}
\end{equation}
where \( \mathbb{I}(g_{ij}, \gamma(t)) = 1 \) if grid cell \( g_{ij} \) has been visited by time \( t \), and 0 otherwise.

The ocean environment is time-varying, with wind and current fields evolving periodically and only partially observable. To handle this, planning is performed in discrete phases over finite time horizons. Overlapping visits to previously covered grids are allowed, reflecting practical needs for data consistency. However, such redundancy does not yield additional reward in our formulation.

The optimization problem is then to find the minimum time \( T^* \) such that the cumulative coverage exceeds a predefined threshold \( \eta \in (0,1) \):
\begin{equation}
T^* = \arg\min_{T} \left\{ T \,\big|\, \mathbb{A}(\gamma(T)) \geq \eta \sum_{g_{ij} \in \mathbb{G}} d_{\text{obs}} \right\}.
\label{eq:t_min}
\end{equation}
Here, \( \sum_{g_{ij} \in \mathbb{G}} d_{\text{obs}} \) denotes the total observable area, and \( \eta \) specifies the required coverage proportion.

\section{The Spatiotemporal Optimization Method}

\subsection{Overview}

As previously discussed, the key challenge in coverage planning for autonomous sailboats lies in the anisotropic performance and passability induced by the uncertainty of time-varying ocean. The conventional CPP methods are insufficient, since they typically assume isotropic traversal capabilities and slowly varying conditions.

To address these issues, we propose a dual-constrained framework that incorporates both morphological constraints in the spatial domain and look-ahead planning in the temporal domain. The core design principles are as follows:

\subsubsection{Backtrack-avoidance morphological constraints}

Decomposition-based methods combined with boustrophedon sweeping are widely used in coverage planning due to their inherent ability to guarantee complete coverage without omission or redundancy. This property is particularly valuable, even when local trajectories are suboptimal, global completeness can still be achieved without requiring backtracking. However, in sailboat-based coverage, such a property does not hold. Predefined sweeping patterns may become infeasible or highly inefficient due to the strong dependence of action feasibility and efficiency on local wind and current conditions.

Thus, we directly regulate the evolving shape of the covered region. Specifically, we suppress fragmented, narrow, or isolated areas that may require costly backtracking, promoting spatial compactness to minimize inefficiencies. This is achieved using perimeter-to-area or convexity-based metrics evaluated within each forecast interval.

\subsubsection{Forecast-aware look-ahead planner}
    
In real-world deployments, sailboats often lack the onboard sensing capabilities to perceive large-scale wind and current fields, and generally lack the computational capacity or model fidelity to predict future environmental changes. As a result, planning often becomes myopic, with the sailboat focusing on short-term rewards while overlooking the long-term implications of environmental trends. Fortunately, in practice, periodically released weather forecasts can serve as a rough priori on future environmental evolution. Although forecasts contain uncertainties, they provide valuable directional information. 
    
To exploit the predictive value of such forecasts, we adopt a look-ahead planning strategy: at each decision point, the algorithm plans over moving time horizons that extends beyond the current environmental state, leveraging forecasted trends to balance immediate efficiency with long-term coverage benefits.

\subsection{Spatiotemporal MCTS planning framework}

Algorithm~\ref{alg:phase_mcts} presents a phase-wise MCTS strategy for planning adaptive action sequences under time-varying environmental conditions. Each planning phase corresponds to a fixed interval \( \Delta t \), during which the wind and current fields are assumed to remain constant. The inputs to the algorithm include:

\begin{itemize}
    \item \( \textit{env}_0 \): the current environment state at the beginning of the phase,
    \item \( \{(\mathbf{W}_{t_0 + k\Delta t}, \mathbf{C}_{t_0 + k\Delta t})\}_{k=0}^K \): the sequence of wind and current fields including the current and \(K \) forecasted intervals,
    \item \( \Delta t \): the planning horizon,
    \item \( N_{\text{iter}} \): the number of MCTS iterations per planning phase.
\end{itemize}

At the beginning of each phase, the environment is cloned and the root node of the tree is initialized. The search then proceeds with \( N_{\text{iter}} \) iterations, each consisting of the following four standard steps\cite{coulom2006efficient}:

\begin{enumerate}
    \item \textbf{Selection}: A node is recursively selected from the root to a child node using the upper confidence bound (UCB) criterion \cite{kocsis2006bandit}:
    \begin{equation}
    \text{UCB}(n) = \bar{S}(n) + C\sqrt{\frac{\log N_{\text{parent}}}{N_n}},
    \label{eq:ucb}
    \end{equation}
    where \( \bar{S}(n) \) is the average score of node \( n \), \( N_n \) is the number of visits to \( n \), and \( N_{\text{parent}} \) is the number of visits to its parent. \( C \) is a positive exploration constant that balances exploitation and exploration.

    \item \textbf{Expansion}: If node \( n \) is expandable (i.e., coverage is incomplete, the horizon is not yet reached, and no dead-end is encountered), an action \( a \in A_{ij} \) is sampled from the feasible (as introduced in the action definition) set at grid \( g_{ij} \), with sampling probability proportional to its heuristic score \( H(a) \) as Eq.~\eqref{eq:heuristic}. A connectivity constraint is applied to avoid creating disconnected uncovered regions (see heuristic score and rollout reward subsection). The selected action is applied to a cloned environment to create a new child node.

    \item \textbf{Simulation}: A rollout is performed from the newly expanded node under the current and K forecasted intervals field sequence \( \{(\mathbf{W}_{t_0 + k\Delta t}, \mathbf{C}_{t_0 + k\Delta t})\} \). The resulting score \( S \) and trajectory \( \tilde{\gamma} \) are returned (see Algorithm~\ref{alg:rollout_strategy}).

    \item \textbf{Backpropagation}: The score \( S \) is propagated along the path from the child node to the root, updating each node’s visit count and average score.
\end{enumerate}

After all iterations, the child of the root node with the highest score is selected. Its corresponding action is executed in the real environment, appended to the output sequence \( \gamma \), and the subtree rooted at this node is promoted as the new root. The process repeats until the current planning phase ends or \( N_{\text{iter}} \) is reached.

\begin{algorithm}[t]
\DontPrintSemicolon
\SetAlgoLined
\caption{\textsc{Phase-wise Planning}}
\label{alg:phase_mcts}

\KwIn{\( \textit{env}_0,\ \{(\mathbf{W}_{t_0 + k\Delta t}, \mathbf{C}_{t_0 + k\Delta t})\}_{k=0}^K,\ \Delta t,\ N_{\text{iter}} \)}
\KwOut{\( \gamma \)}

Initialize environment: \( \textit{env} \gets \textit{env}_0 \)\;
Set initial time: \( t_0 \gets \textit{env}.t \)\;
Initialize action sequence: \( \gamma \gets [\ ] \)\;
Initialize MCTS tree with root node storing \( \textit{env} \)\;

\While{\( \textit{env} \) not terminated \textbf{and} \( \textit{env}.t < t_0 + \Delta t \)}{
    \For{\( i = 1 \) \KwTo \( N_{\text{iter}} \)}{
        \tcp{Selection}
        Select node \(n\) via \text{UCB}(\( n \));

        \tcp{Expansion}
        \If{\( n \) is expandable}{
            Retrieve feasible action set\( A_{ij} \), at grid \( g_{ij} \)\;
            Sample \( a \in A_{ij} \) with probability \( \propto H(a) \)\;
            Subject to connectivity constraint\;

            Clone environment at node \( n \): \( \textit{env}_n \)\;
            Apply \( a \) in \( \textit{env}_n \) to generate next state\;
            Create child node \( n_{\text{new}} \) storing \( \textit{env}_n \)\;
            Add \( n_{\text{new}} \) to \( n \); update \( n \gets n_{\text{new}} \)\;
        }

        \tcp{Simulation}
        Clone environment at node \( n \): \( \textit{env}_n \)\;
        \( (S, \tilde{\gamma}) \gets \textsc{Rollout}(\textit{env}_n,\ \{(\mathbf{W}_{t_0 + k\Delta t}, \mathbf{C}_{t_0 + k\Delta t})\}_{k=0}^{K}) \)\;

        \tcp{Backpropagation}
        Propagate score \( S \) from \( n \) to root\;
        Update \( N_n \), \( \bar{S}(n) \), and \( N_{\text{parent}} \) along path\;
    }

    \tcp{Action Execution}
    Select child of root with highest score as \( a \)\;
    Apply \( a \) in \( \textit{env} \), update state\;
    Append \( a \) to \( \gamma \)\;
    Promote subtree rooted at \( a \) as new root\;
}
\Return \( \gamma \)\;
\end{algorithm}

Algorithm~\ref{alg:rollout_strategy} outlines the Rollout Simulation procedure used during the \textbf{Simulation} step of MCTS. Starting from a cloned environment \(\textit{env}_0\), the algorithm simulates a sequence of actions over a horizon of \(K\Delta t\), using both the current and forecasted wind/current fields \(\{(\mathbf{W}_{t_0 + k\Delta t}, \mathbf{C}_{t_0 + k\Delta t})\}_{k=0}^K\) to approximate future dynamics.

At each simulation step, the algorithm identifies the current grid cell \( g_{ij} \) and selects the corresponding field values \( \{(\mathbf{W}, \mathbf{C})\} \) according to the current time \( \textit{env}.t \). The feasible action set \( A_{ij} \subseteq A \) is then retrieved. Each action \( a \in A_{ij} \) is assigned a heuristic score \( H(a) \) as Eq.~\eqref{eq:heuristic}, and one action is sampled using an \( \varepsilon \)-greedy policy proportional to these scores.

To ensure coverage consistency, a connectivity constraint is enforced to prevent the split of uncoverd area large than larger than threshold \( A_{\text{thres}} \). The selected action is then applied to update the environment, and the process iterates until the K time horizon ends or termination is triggered. Finally, the rollout score \( S = {R}(\tilde{\gamma}) \) is computed based on the resulting trajectory \( \tilde{\gamma} \) and returned for use in the backpropagation step.

\begin{algorithm}[t]
\DontPrintSemicolon
\SetAlgoLined
\caption{\textsc{Rollout}}
\label{alg:rollout_strategy}

\KwIn{\( \textit{env}_0, \{(\mathbf{W}_{t_0 + k\Delta t}, \mathbf{C}_{t_0 + k\Delta t})\}_{k=0}^{K} \)}
\KwOut{\( S = {R}(\tilde{\gamma}),\ \tilde{\gamma} \)}

Initialize environment: \( \textit{env} \gets \textit{env}_0 \)\;
Initialize action sequence: \( \tilde{\gamma} \gets [\ ] \)\;

\While{\( \textit{env}.t < t_0 + K\Delta t \) \KwAnd not terminated}{
    Get current cell: \( g_{ij} \gets \textit{env}.\mathrm{Position} \)\;
    Determine field \( \{(\mathbf{W}, \mathbf{C}) \}\) at \( \textit{env}.t \) from field sequence\;
    Retrieve feasible actions: \( A_{ij} \subseteq A \)\;

    \ForEach{\( a \in A_{ij} \)}{
        Compute heuristic score \( H(a) \)\;
    }

    Sample \( a \in A_{ij} \) using \( \varepsilon \)-greedy based on \( H(a) \)\;

    \If{\( a \) violates connectivity constraint}{
        Reject \( a \), re-sample\;
    }

    Apply \( a \) to \( \textit{env} \), update internal state\;
    Append \( a \) to \( \tilde{\gamma} \)\;
}

Compute rollout score: \( S \gets {R}(\tilde{\gamma}) \)\;
\Return \( (S, \tilde{\gamma}) \)\;
\end{algorithm}

\subsection{Heuristic score and rollout reward}

To efficiently identify promising actions within a large action space, each candidate action \( a \in A_{ij} \) is scored by the heuristic function \( H(a) \), which jointly considers three factors: net coverage efficiency, spatial regularity, and positional preference. During node expansion, an $\varepsilon$-greedy strategy is used. With probability $\varepsilon$, an action is selected uniformly at random; with probability $1-\varepsilon$, an action is sampled proportionally to its heuristic score $H(a)$. 

\begin{equation}
H(a) = \mathrm{efficiency_{\text{net}}}(a) \cdot \mathrm{regularity}(a) \cdot \mathrm{position}(a)
\label{eq:heuristic}
\end{equation}

The efficiency term quantifies the coverage gain per unit time, given by
\begin{equation}
\mathrm{efficiency_{\text{net}}}(a) = \frac{\Delta A(a)}{T(a)}.
\label{eq:efficiency}
\end{equation}
where \( \Delta A(a) \) is the number of newly covered pixels, and \( T(a) \) is the estimated execution time of action \( a \) under current wind and current conditions.

\begin{figure}[!t]
\centering
\scalebox{0.75}{
\includegraphics[width=\linewidth]{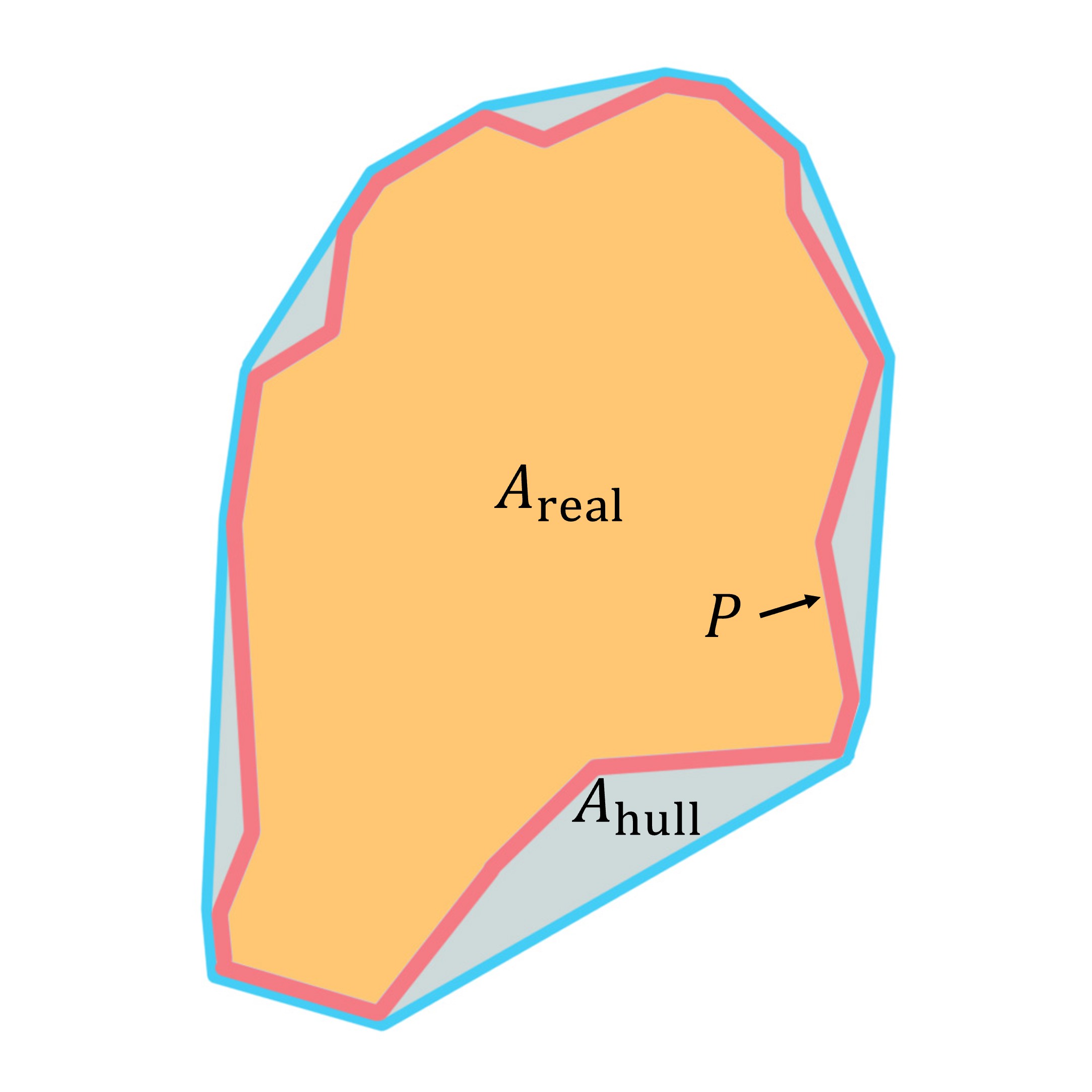}
}
\caption{Key parameters in regularity term}
\label{FIG:Hull}
\end{figure}

The regularity term, illustrated in Fig.~\ref{FIG:Hull}, measures the morphological compactness of the updated coverage map as
\begin{equation}
\mathrm{regularity}(a) = S^{\text{cov}}_{\text{convex}} \cdot S^{\text{uncov}}_{\text{convex}} \cdot S^{\text{cov}}_{\text{shape}} \cdot S^{\text{uncov}}_{\text{shape}}.
\label{eq:regularity}
\end{equation}

Only the largest connected components of the covered and uncovered regions are considered, as the planner prohibits splitting the uncovered region into multiple large disconnected parts. The convexity score \( S_{\text{convex}} \) is computed by
\begin{equation}
S_{\text{convex}} = \frac{A_{\text{real}}}{A_{\text{hull}}},
\label{eq:convexity}
\end{equation}
where \( A_{\text{real}} \) is the actual area and \( A_{\text{hull}} \) is the area of its convex hull. The shape score \( S_{\text{shape}} \), defined in Eq.~\eqref{eq:shape}, measures how close a region is to a circle:
\begin{equation}
S_{\text{shape}} = \frac{4\pi A_{\text{real}}}{P^2},
\label{eq:shape}
\end{equation}
where \( A_{\text{real}} \) and \( P \) denotes the area and perimeter,respectively. 

To encourage early expansion toward peripheral regions, we assign each grid cell \( g_{ij} \in \mathbb{G} \) a spatial score based on its Euclidean distance from the map center \( \mathbf{c} \):

\begin{equation}
d_{ij} = \left\| \mathbf{g}_{ij} - \mathbf{c} \right\|_2.
\label{eq:distance}
\end{equation}

These distances are then normalized to form a spatial weighting distribution over the entire map:

\begin{equation}
w_{ij} = \frac{d_{ij}}{\sum_{g_{ij} \in \mathbb{G}} d_{ij}},
\label{eq:weight}
\end{equation}

such that
\(
\sum_{g_{ij} \in \mathbb{G}} w_{ij} = 1
\). This normalization preserves the relative differences between grid distances and ensures that cells farther from the center receive proportionally higher weights. 

For a given action \( a \), let \( g_{\text{end}}(a) \in  \mathbb{G} \) denote the terminal grid cell reached by executing \( a \). The position score is defined as

\begin{equation}
\mathrm{position}(a) = d_{g_{\text{end}}(a)} \cdot w_{g_{\text{end}}(a)}.
\label{eq:position}
\end{equation}

Eq.~\eqref{eq:position} integrates both absolute distance (Eq.~\eqref{eq:distance}) and normalized spatial importance (Eq.~\eqref{eq:weight}) to favor early coverage of peripheral regions.

After simulating an action sequence \( \tilde{\gamma} \) until the predefined planning horizon \( t + K \Delta t \) is reached, the rollout evaluation computes a cumulative reward score \({R}(\tilde{\gamma}) \), which guides the backpropagation step in MCTS. Unlike a single-snapshot estimate, the score aggregates multiple stages to balance coverage efficiency and spatial regularity throughout the forecast horizon, as defined in Eq.~\ref{eq:rollout_reward}

\begin{equation}
{R}(\tilde{\gamma}) = \sum_{k=1}^{K} \beta^{k-1} \Big( \mathrm{regularity}(\tilde{\gamma}_k) \cdot \mathrm{efficiency_{\text{pen}}}(\tilde{\gamma}_k) \Big),
\label{eq:rollout_reward}
\end{equation}
where \( \tilde{\gamma}_k \) denotes the trajectory up to stage \( k \), and 
\( \mathrm{regularity}(\tilde{\gamma}_k) \) follows the same formulation as Eq.~\ref{eq:regularity}, 
evaluating the convexity- and shape-based compactness of the covered region. 
The discount factor $\beta \in (0, 1]$ gradually downweights rewards from later stages 
to emphasize near-term performance. At each stage, the term 
\( \mathrm{efficiency}_{\text{pen}}(\tilde{\gamma}_k) \) reflects the penalized coverage 
efficiency, which encourages uniform exploration while penalizing redundant visits. 
It is defined as the ratio between a redundancy-aware coverage score and the total sailing 
time for that segment as Eq.~\eqref{eq:efficiency2}.

\begin{equation}
\mathrm{efficiency_{\text{pen}}}(\tilde{\gamma}_k) = \frac{U(\tilde{\gamma}_k)}{T(\tilde{\gamma}_k)}
\label{eq:efficiency2}
\end{equation}

The numerator \( U(\tilde{\gamma}_k) \) quantifies the spatial coverage quality while accounting for redundancy penalties. For each grid cell \( g_{ij} \in \mathbb{G} \), let \( \mathcal{M}_{ij} \) denote its pixel-level coverage matrix and \( N_{ij} \) its total pixel count. The normalized coverage score is computed as 

\begin{equation}
U(\tilde{\gamma}_k) = \frac{1}{|\mathbb{G}|} \sum_{g_{ij} \in \mathbb{G}} \Bigg( \frac{|\mathcal{M}_{ij} > 0|}{N_{ij}} - \alpha \sum_{c > 1} \Big( \frac{|\mathcal{M}_{ij} = c|}{N_{ij}} (c - 1) \Big) \Bigg).
\label{eq:uniformity}
\end{equation}

Here, the first term measures the fraction of pixels covered at least once, while the second term penalizes repeated coverage. For each unique coverage count \( c > 1 \), the fraction of pixels covered exactly \( c \) times is weighted by its repetition degree \( c - 1 \) and scaled by a penalty factor \( \alpha \in [0, 1] \). The score is averaged over all valid grid cells.

This formulation ensures that the efficiency term favors trajectories that maximize broad single-pass coverage while minimizing excessive revisits at each stage. When combined with the total sailing time \( T(\tilde{\gamma}_k) \), it yields a time-normalized evaluation of spatial uniformity.

The term \( \mathrm{regularity}(\tilde{\gamma}_k) \) evaluates the spatial compactness and shape consistency of the covered region at the end of stage \( k \), based on the same formulation used in the heuristic score (Eq.~\eqref{eq:regularity}), but applied to the environment snapshot at each stage. This staged cumulative reward design encourages the planner to maintain desirable coverage properties consistently across the entire horizon.

\section{Simulation-Based Evaluation}

\subsection{Simulation setup}

To evaluate the effectiveness of the proposed planning method, we conduct comparative simulations in fully randomized inhomogeneous and time-varying ocean environments (each scenario identified by its random seed) across three strategies:

\begin{enumerate}
    \item \textbf{Baseline (Boustrophedon)}: A manually designed zigzag pattern that ignores environmental dynamics.
    \item \textbf{Proposed ($K=0$)}: MCTS-based planning with no forecast horizon (current field only).
    \item \textbf{Proposed ($K=1$)}: MCTS-based planning with a one-step forecast horizon.
\end{enumerate}

The baseline path is manually constructed to ensure full coverage while minimizing maneuvers. Although boustrophedon patterns can be adapted to wind or current fields (e.g., by changing sweeping directions), we fix the path across all runs to provide a fair and stable reference. When strong currents render a scheduled motion physically infeasible, the baseline agent waits at its current location until the action becomes executable. These assumptions are intended to highlight the performance advantages of the proposed adaptive planner under realistic marine dynamics.

All strategies operate under a unified simulation environment and parameter configuration. The simulation map is a $10 \times 10$ grid, where each grid cell represents a $100\,\mathrm{m} \times 100\,\mathrm{m}$ region. The autonomous sailboat starts at the top-left corner \( g_{1,1} \), and the observation radius is fixed at \( d_{\text{obs}} = 72\,\mathrm{m} \). Each planning phase spans \( \Delta t = 300\,\mathrm{s} \).

\subsubsection{Environmental field generation}

The wind and current fields \( \mathbf{W}_t \) and \( \mathbf{C}_t \) are initialized and evolved in a fully randomized manner, generated using a multiscale strategy:

\begin{itemize}
    \item Coarse scale: A $2 \times 2$ grid is initialized with uniform random noise, smoothed by a Gaussian filter, and upsampled to $10 \times 10$.
    \item Fine scale: A $10 \times 10$ grid is separately initialized and smoothed using the same process.
\end{itemize}

The final vector field is computed as a weighted sum (0.6 for coarse, 0.4 for fine) and normalized to physical bounds: flow speed in $[v_{\min}, v_{\max}] = [0.2, 5.0]\,\mathrm{m/s}$ and direction in $[0^\circ, 359^\circ]$. To emulate forecast uncertainty, each predicted field (for $k > 0$) is modeled as the corresponding true field perturbed with bounded random noise. The perturbation grows linearly with the forecast horizon, such that the maximum deviation at step $k$ reaches $(k\epsilon_v, k\epsilon_\theta)$, where $\epsilon_v = 0.05$ and $\epsilon_\theta = 5^\circ$, applied independently at each forecast step.

\subsubsection{Algorithm configuration}

All simulations are executed on an AMD EPYC 7K62 CPU with 48 cores. For each MCTS decision step:

\begin{itemize}
    \item Number of MCTS iterations: \( N_{\text{iter}} = 64 \)
    \item Number of rollout workers: 96 (parallelized)
    \item Rollouts per worker: 3
    \item UCB exploration constant: \( C = 2.5 \)
    \item Random exponent: \( p \sim \mathcal{U}[0.25, 4] \), used in \( \text{regularity}(\tilde{\gamma})^p \) (Eq.~\eqref{eq:heuristic})
\end{itemize}

Each planning step therefore involves more than 18,000 rollouts. The connectivity threshold is set as \( A_{\text{thres}} = 3000\,\mathrm{m}^2 \), which also filters small voids in the compactness evaluation \( S_{\text{shape}} \).

\subsubsection{Heuristic and rollout parameters}

\begin{itemize}
    \item In Eq.~\eqref{eq:position}, the distance term \( d_{ij} \) is square-rooted to smooth position bias.
    \item In Eq.~\eqref{eq:rollout_reward}, \( \mathrm{efficiency}_{\text{pen}}(\tilde{\gamma}_k) \) is squared to emphasize efficient coverage.
    \item In Eq.~\eqref{eq:uniformity}, the redundancy penalty coefficient is set to \( \alpha = 0.2 \).
    \item The $\varepsilon$-greedy exploration probability is set to \( \varepsilon = 0.3 \).
    \item The rollout accumulation discount factor is set to \( \beta = 0.2 \).
\end{itemize}

\subsection{Simulation results}
\begin{figure*}[!t]
\centering

\includegraphics[width=0.31\linewidth]{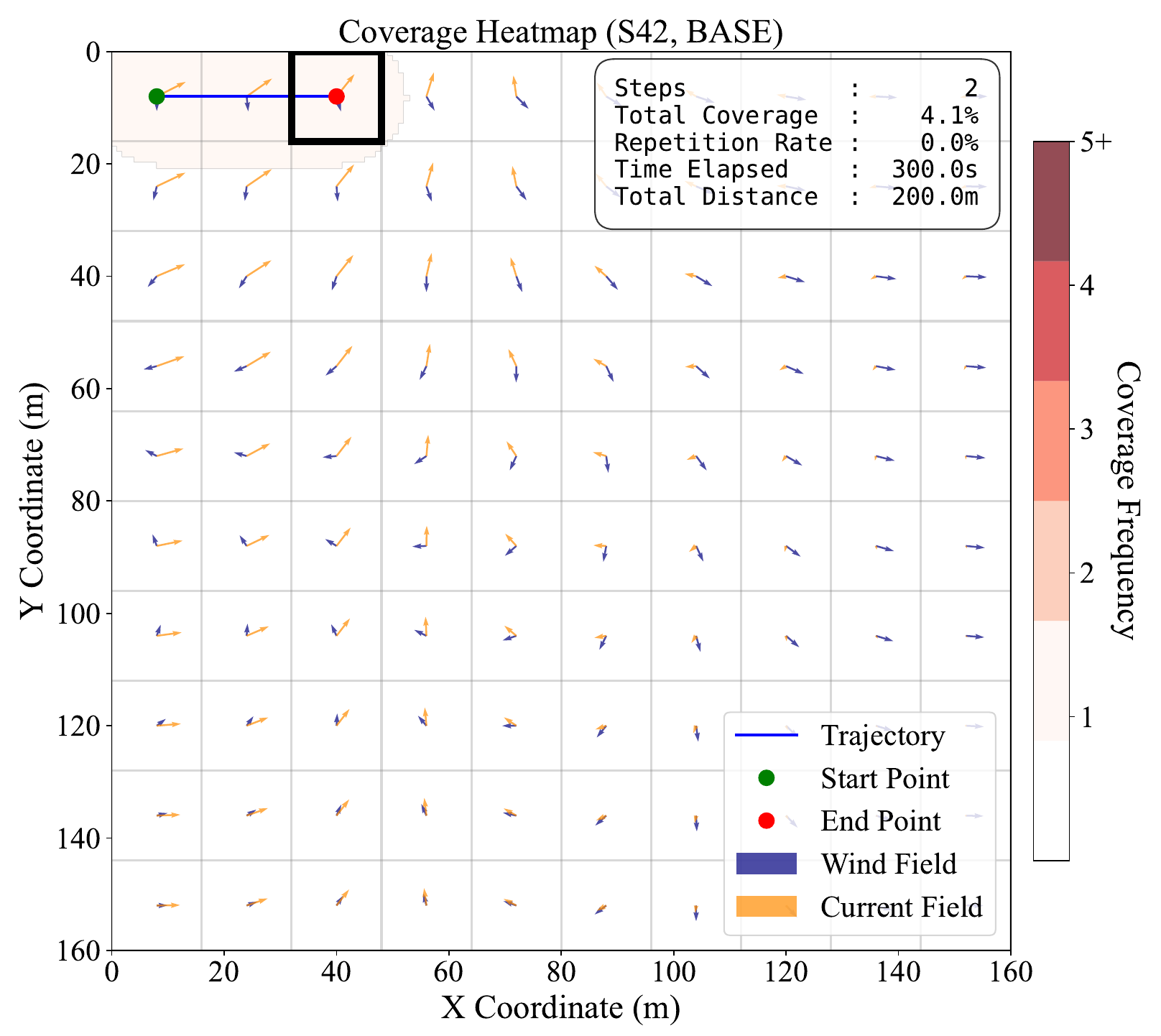}
\hfill
\includegraphics[width=0.31\linewidth]{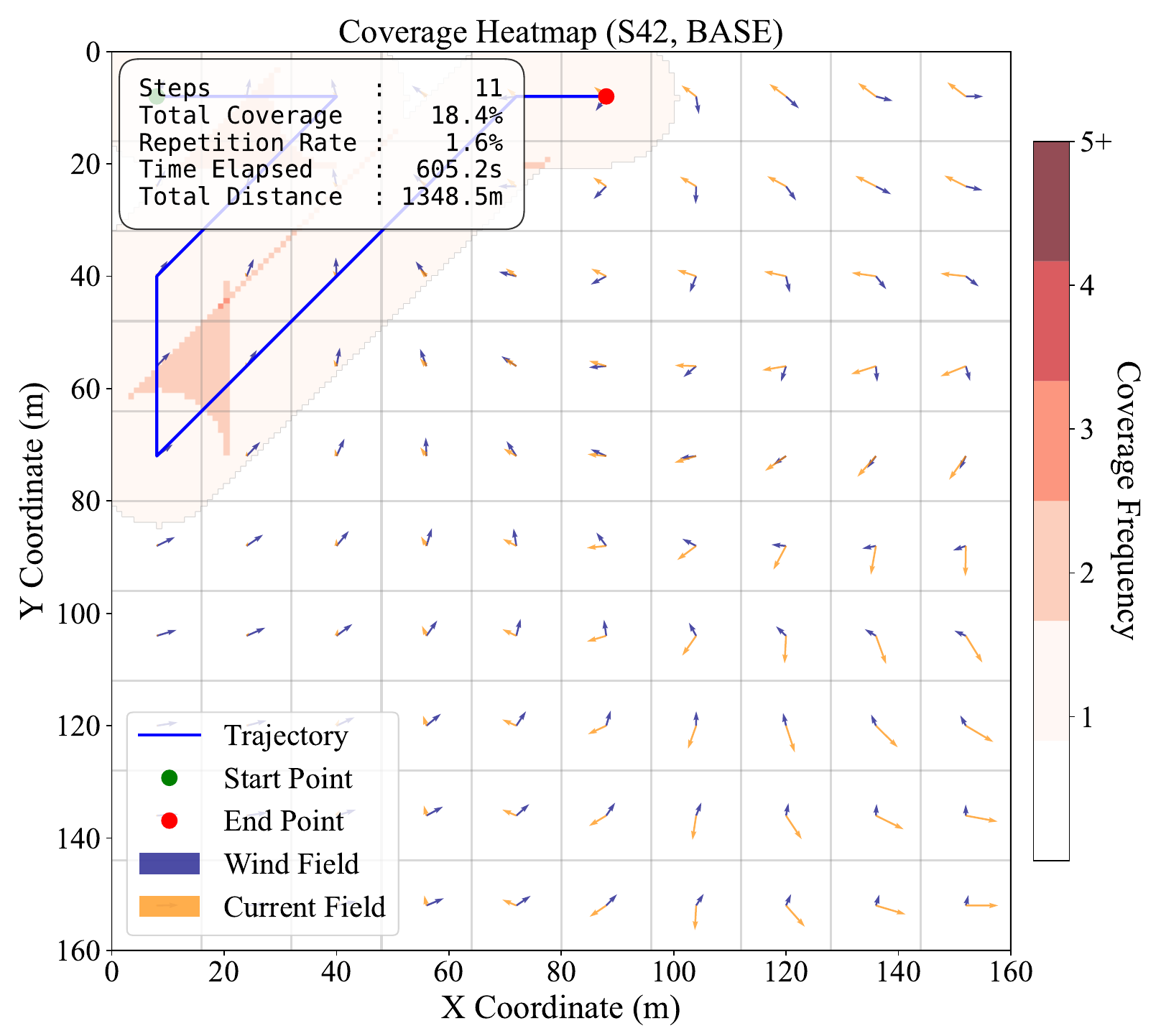}
\hfill
\includegraphics[width=0.31\linewidth]{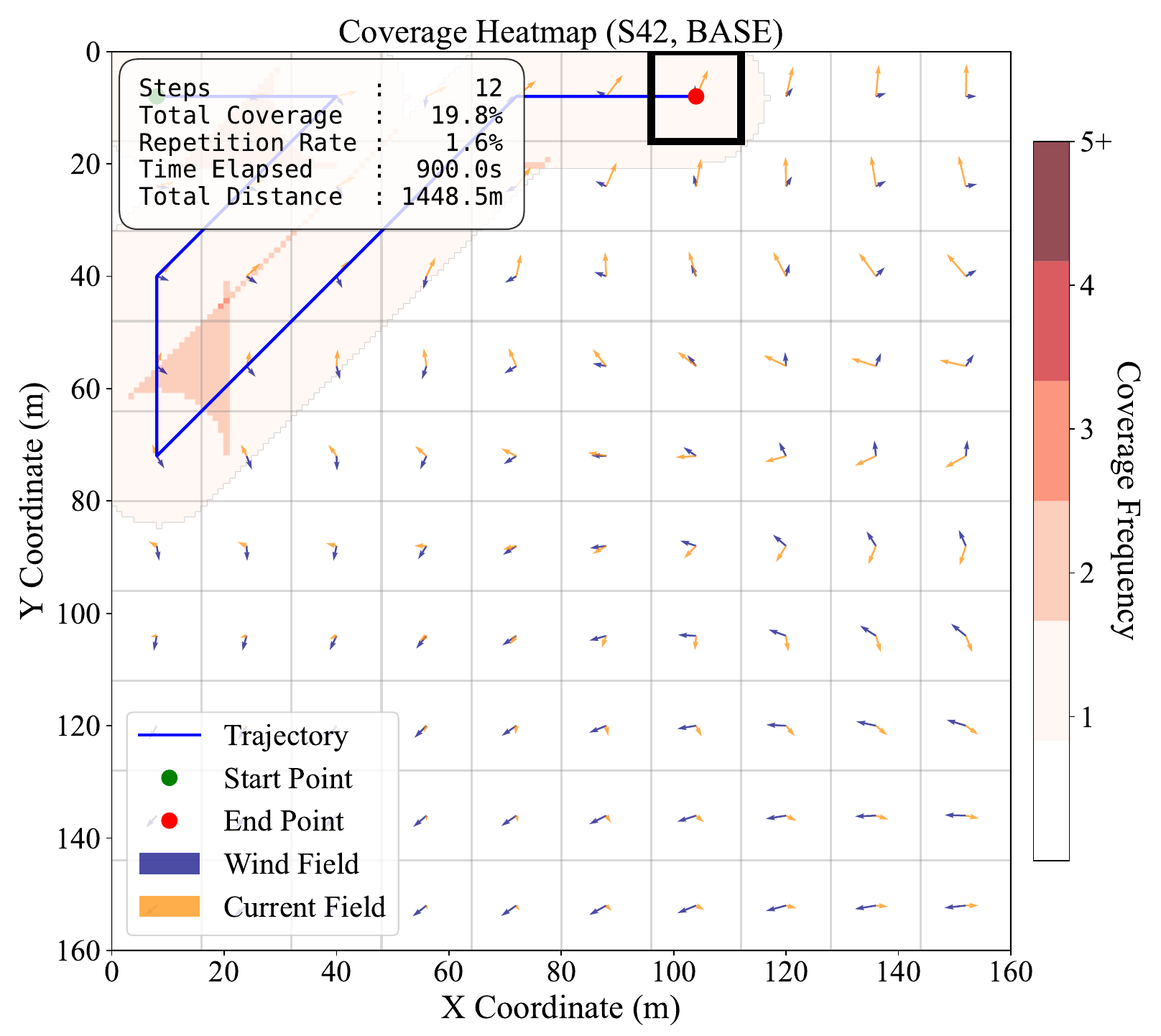}

\vspace{1mm}

\includegraphics[width=0.31\linewidth]{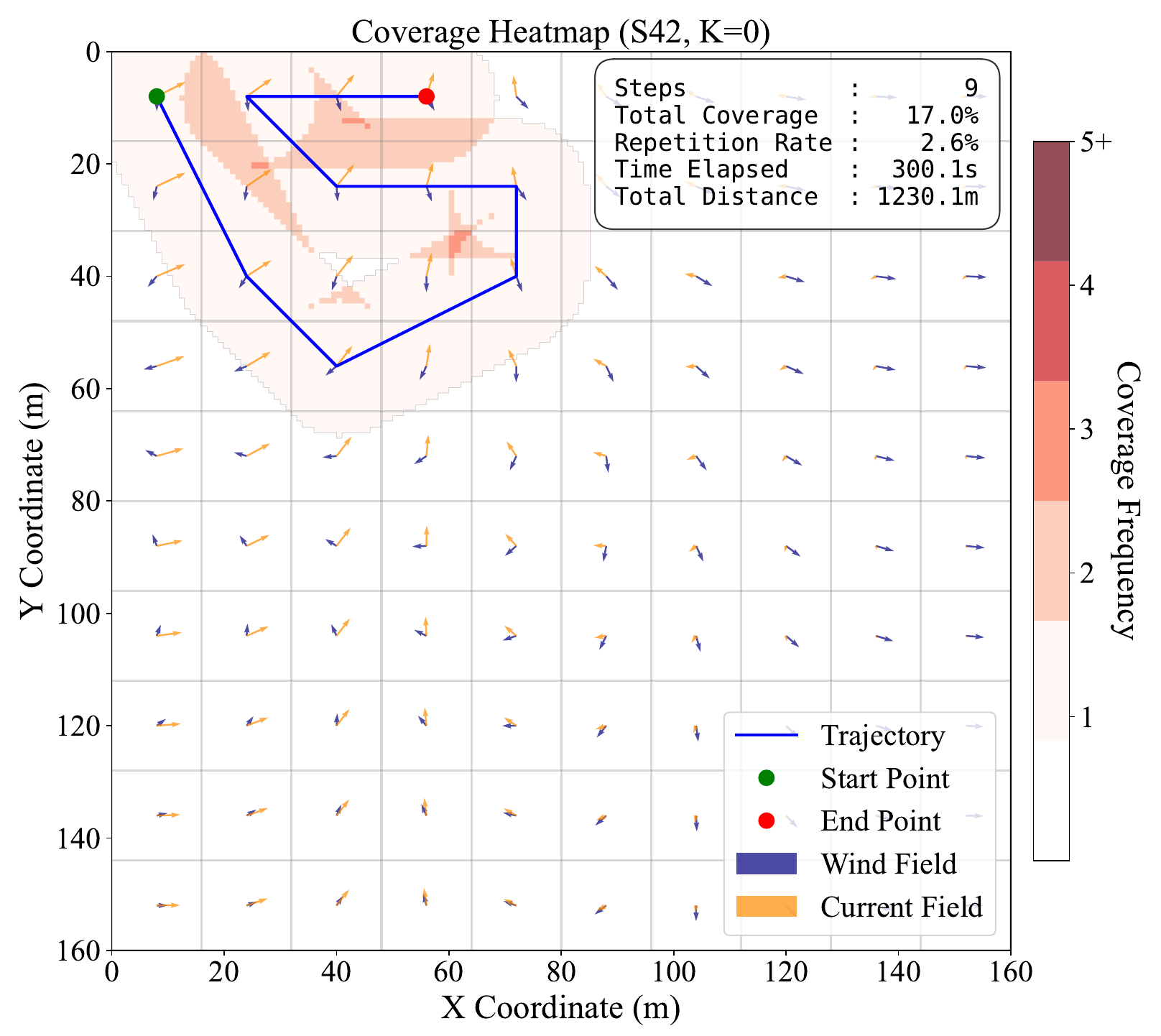}
\hfill
\includegraphics[width=0.31\linewidth]{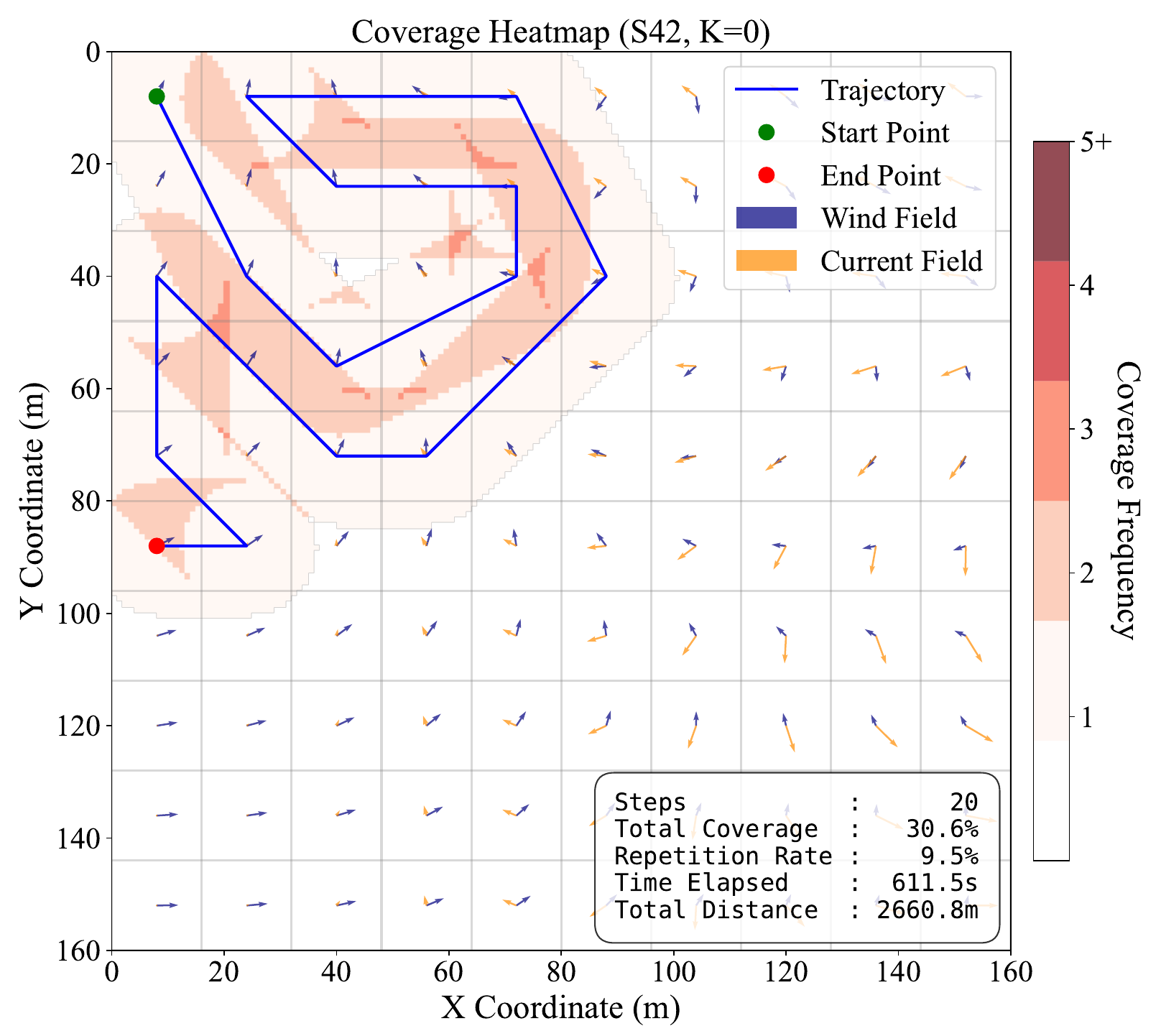}
\hfill
\includegraphics[width=0.31\linewidth]{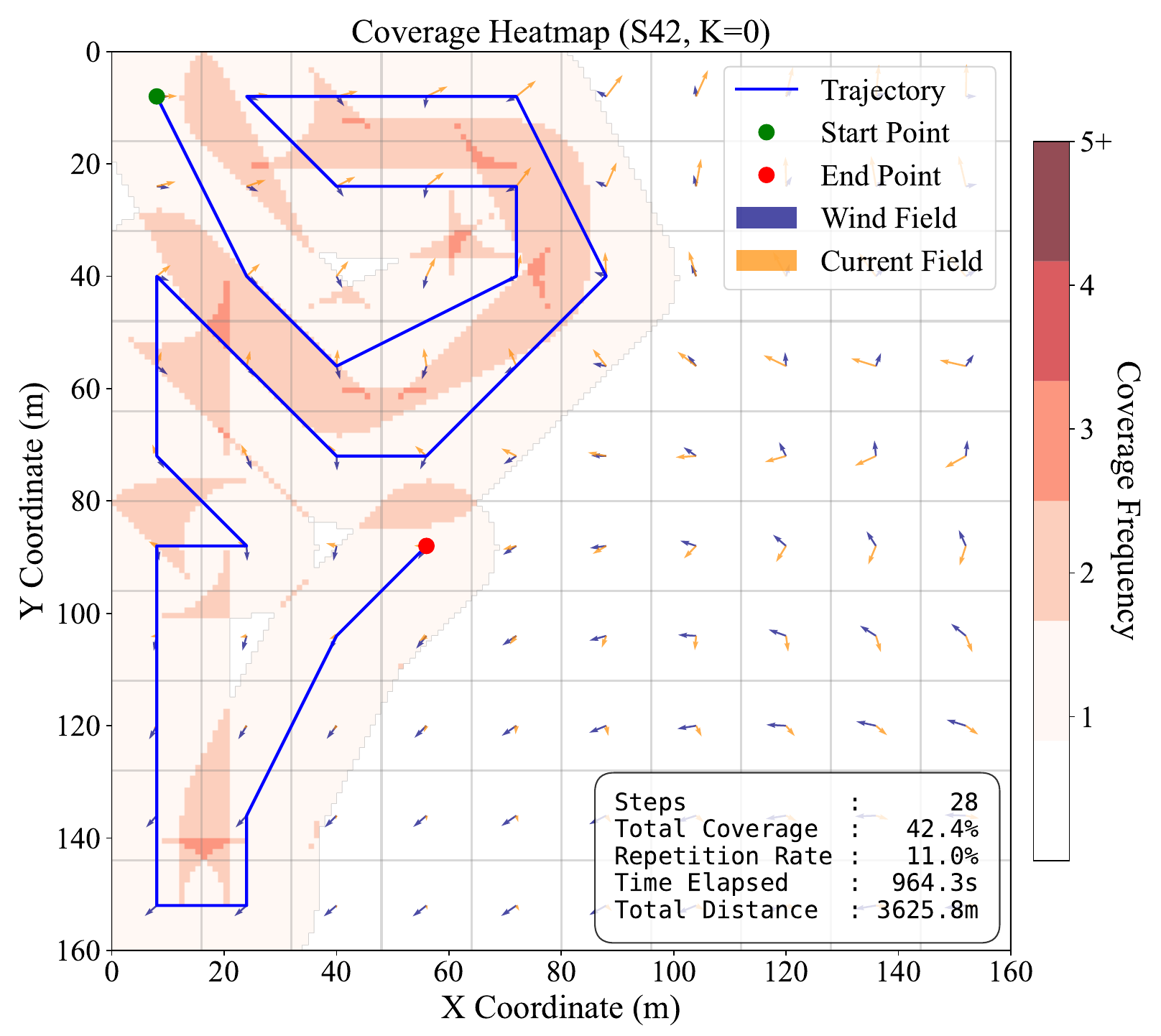}

\vspace{1mm}

\begin{minipage}[t]{0.31\linewidth}
    \centering
    \includegraphics[width=\linewidth]{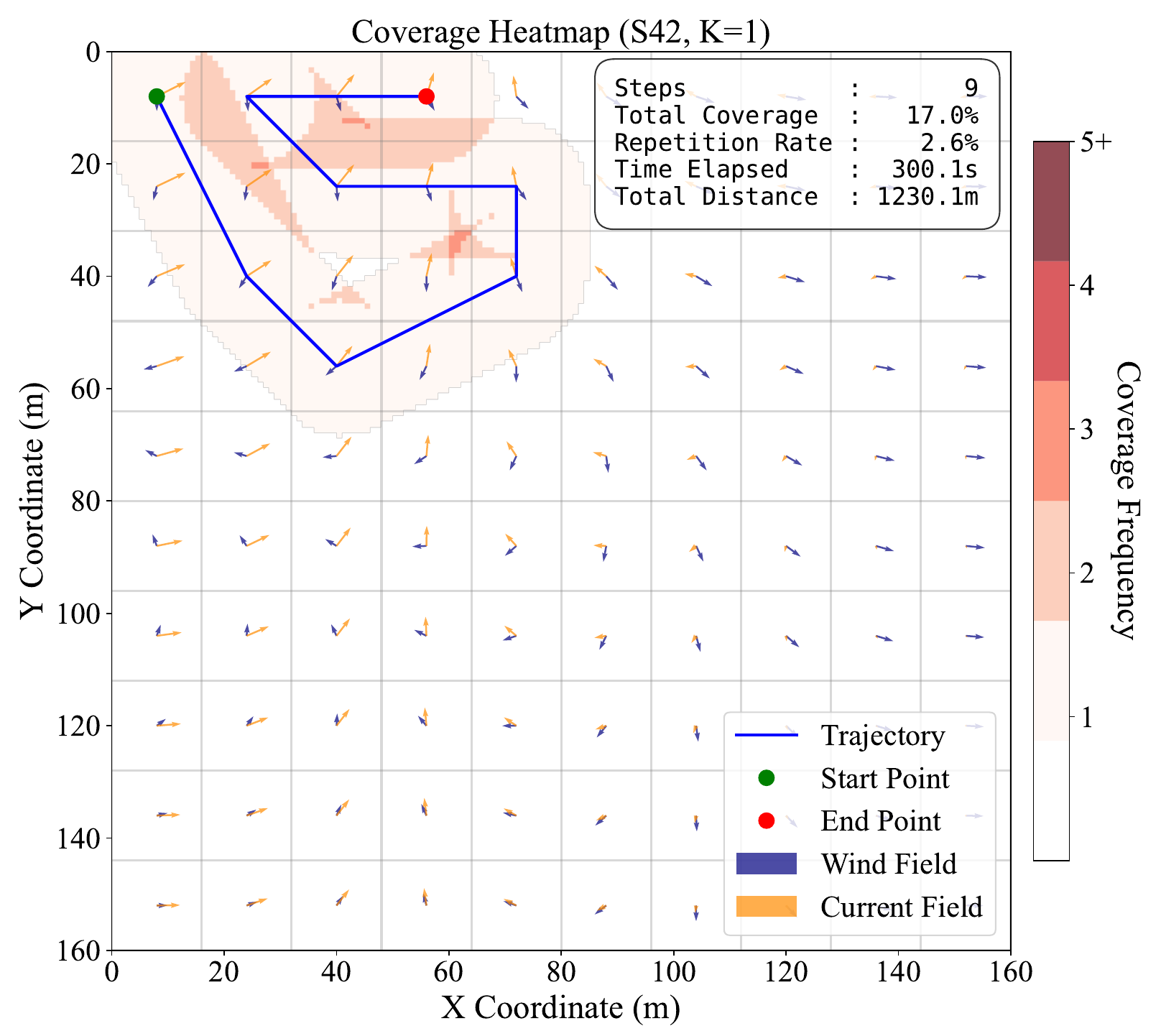}
    
    \vspace{1mm}
    \footnotesize Stage 1
\end{minipage}
\hfill
\begin{minipage}[t]{0.31\linewidth}
    \centering
    \includegraphics[width=\linewidth]{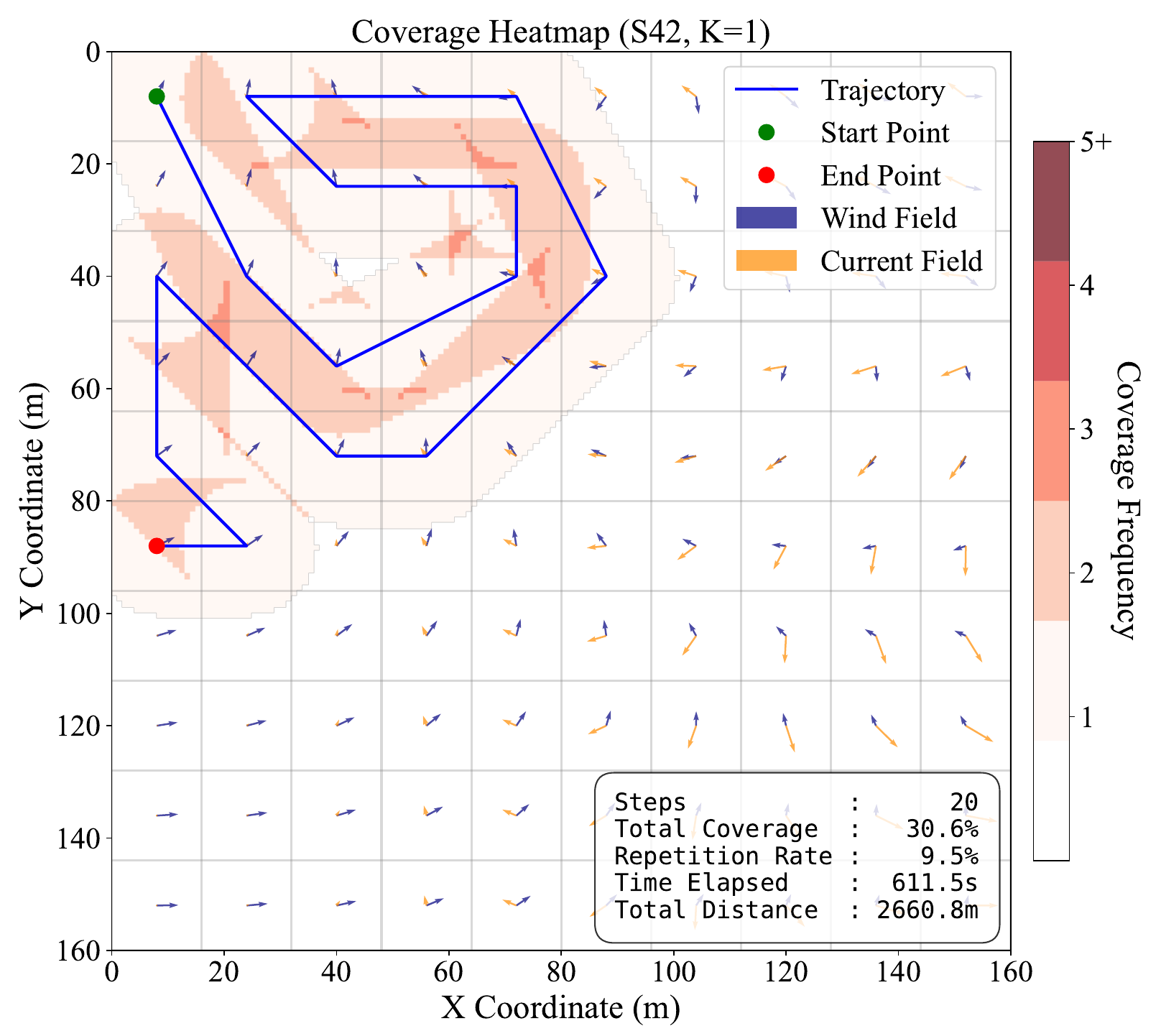}
    
    \vspace{1mm}
    \footnotesize Stage 2
\end{minipage}
\hfill
\begin{minipage}[t]{0.31\linewidth}
    \centering
    \includegraphics[width=\linewidth]{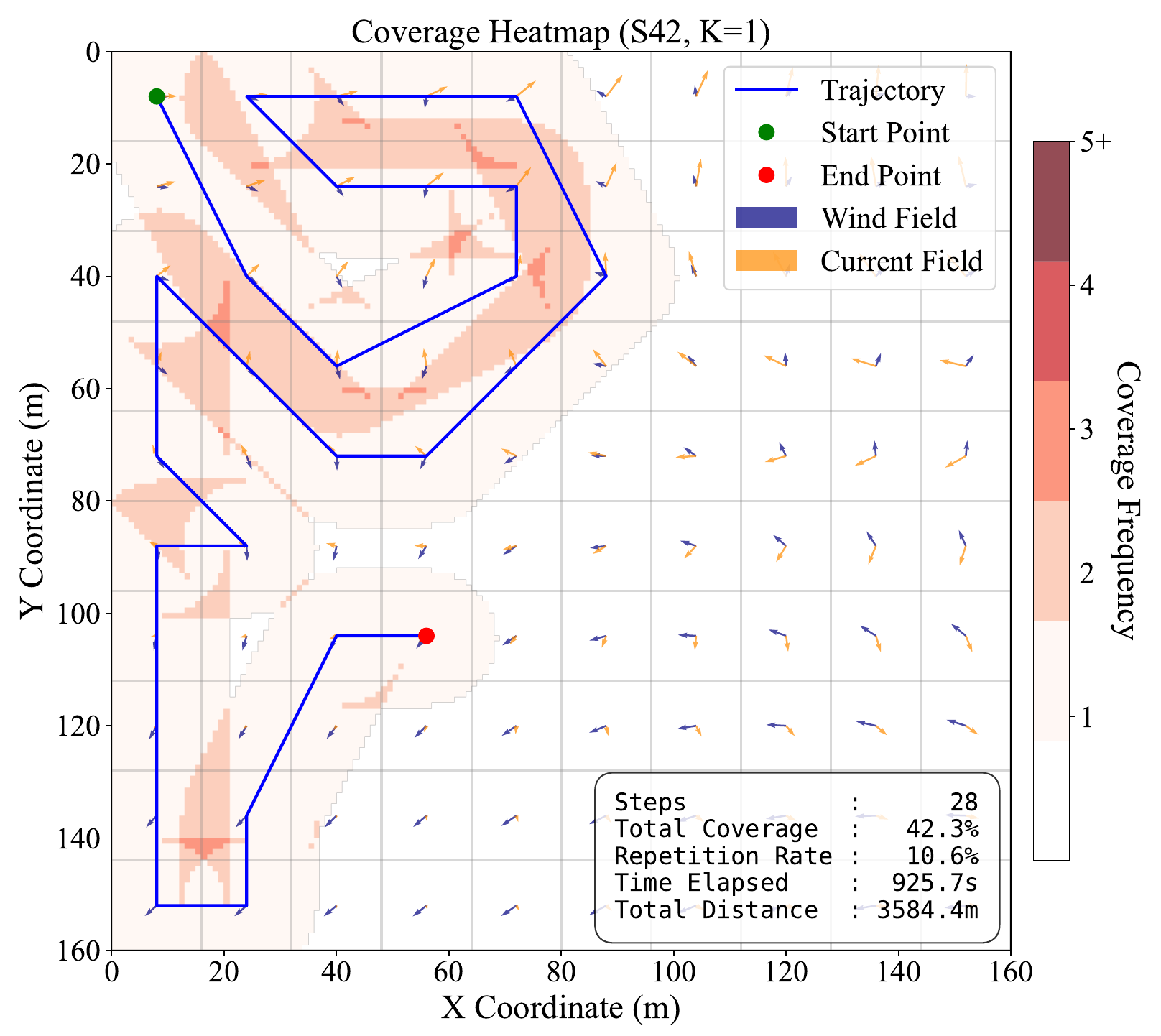}
    
    \vspace{1mm}
    \footnotesize Stage 3
\end{minipage}

\caption{Detailed coverage process of S42 (Stage 1--3). The baseline method stalls when facing temporary inaccessibility (highlighted in black box), while both proposed variants maintain effective coverage. $K{=}0$ and $K{=}1$ exhibit similar results due to shared parameters in early stages.}

\label{FIG:detail1}
\vspace{1mm}
\footnotesize\textit{Note:} The current field uses a scale one-third that of the wind field to highlight the flow intensity. The same scaling is used in all subsequent figures.
\end{figure*}

\begin{figure*}[!t]
\centering

\includegraphics[width=0.31\linewidth]{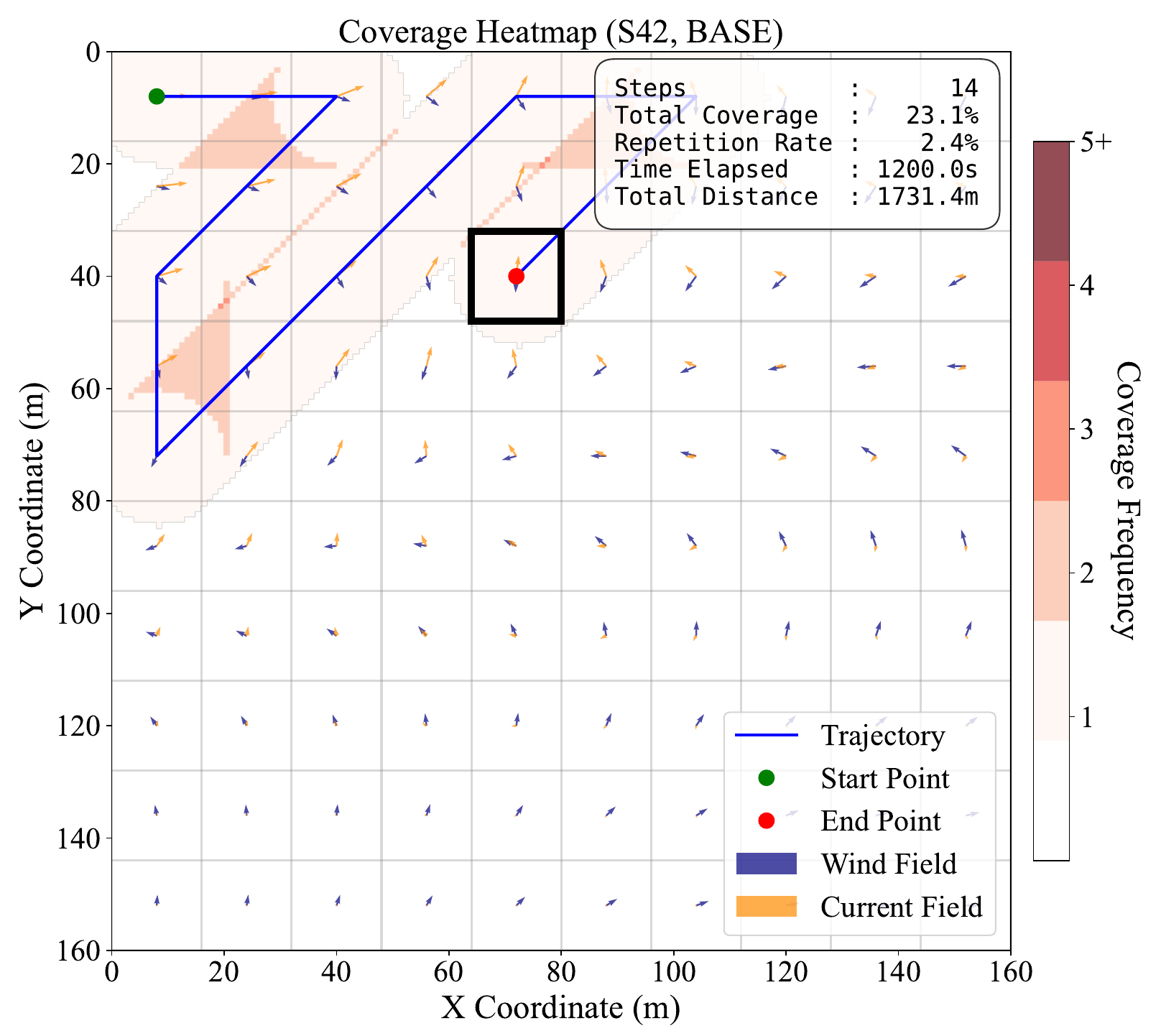}
\hfill
\includegraphics[width=0.31\linewidth]{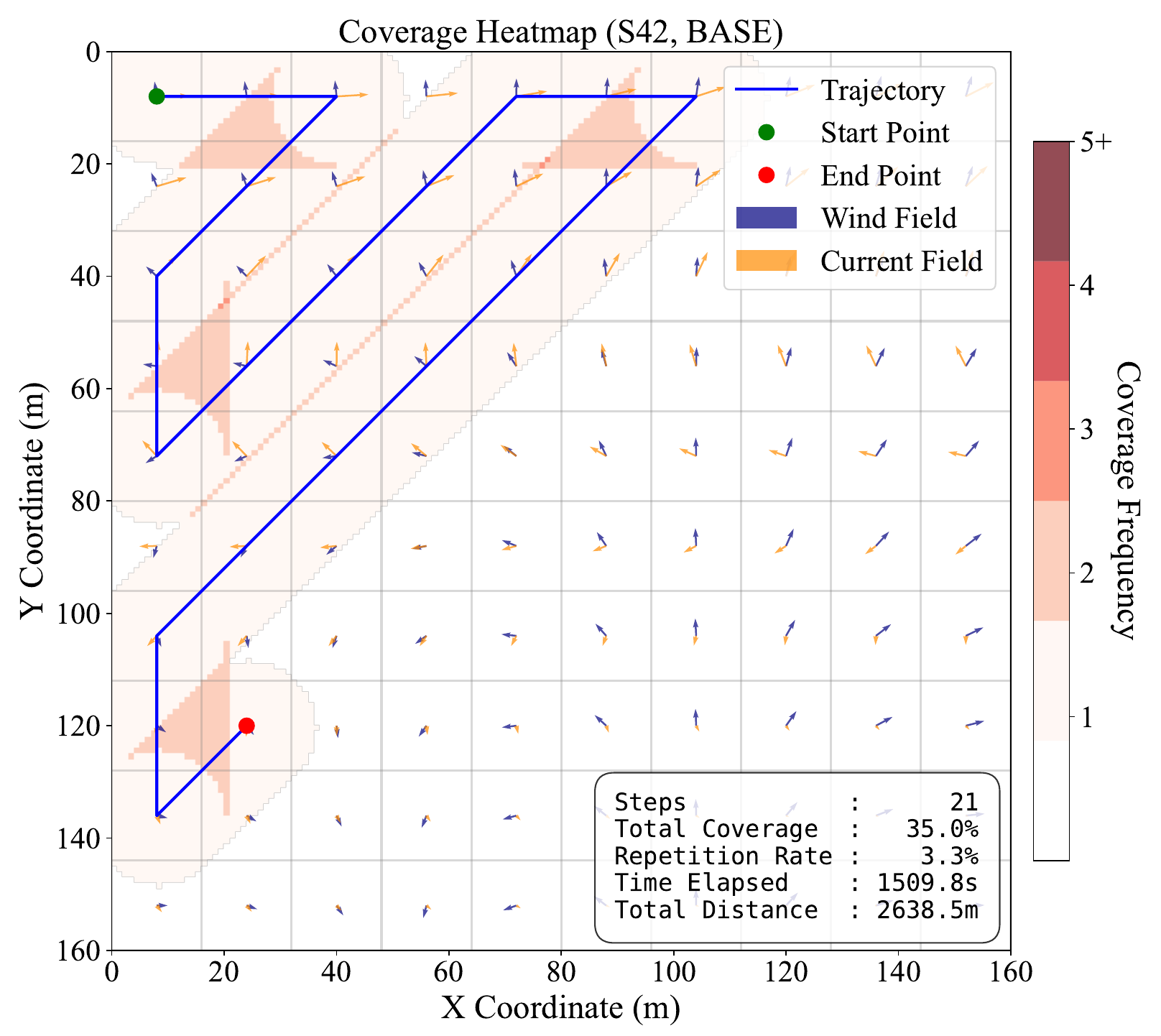}
\hfill
\includegraphics[width=0.31\linewidth]{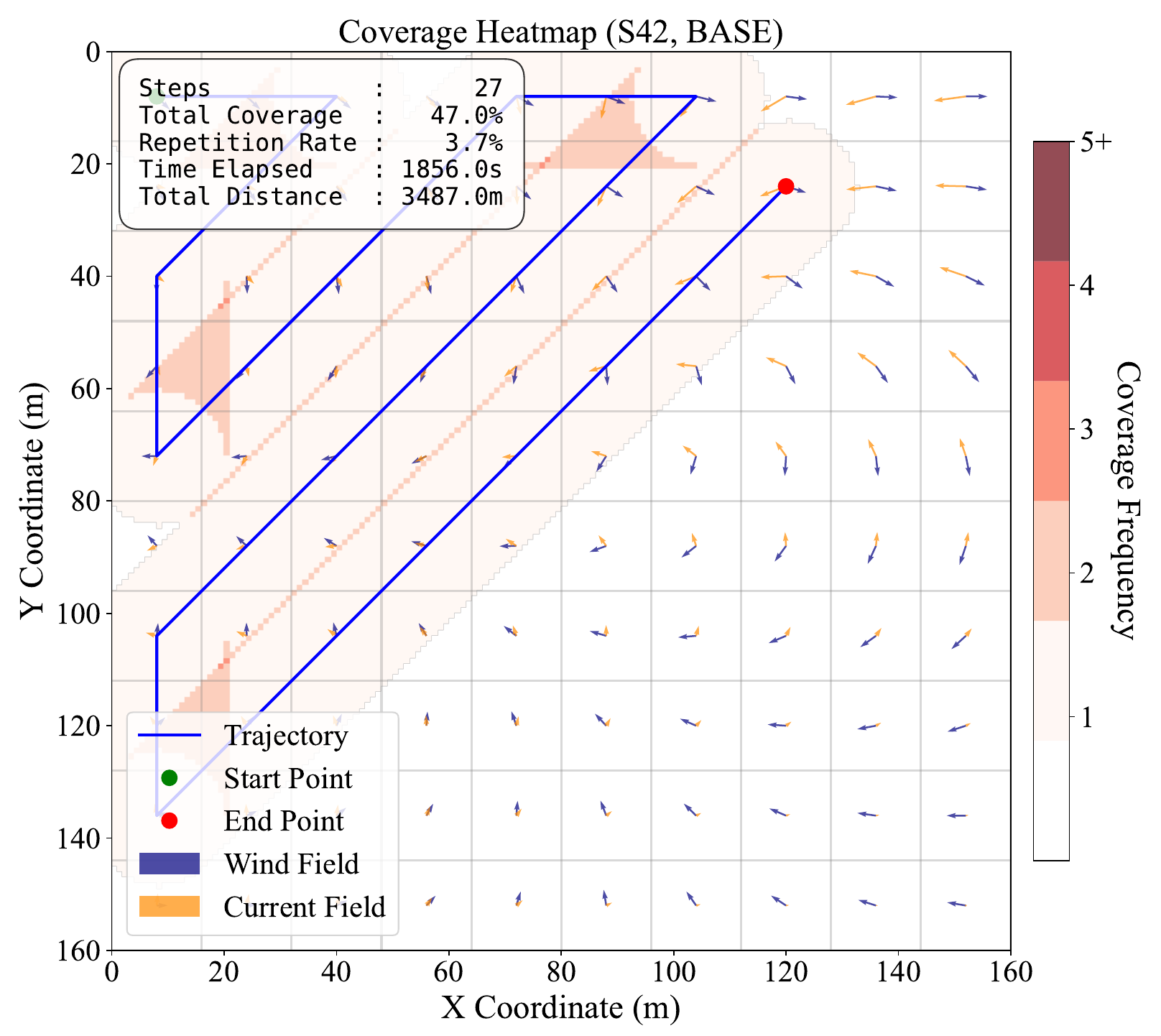}

\vspace{1mm}

\includegraphics[width=0.31\linewidth]{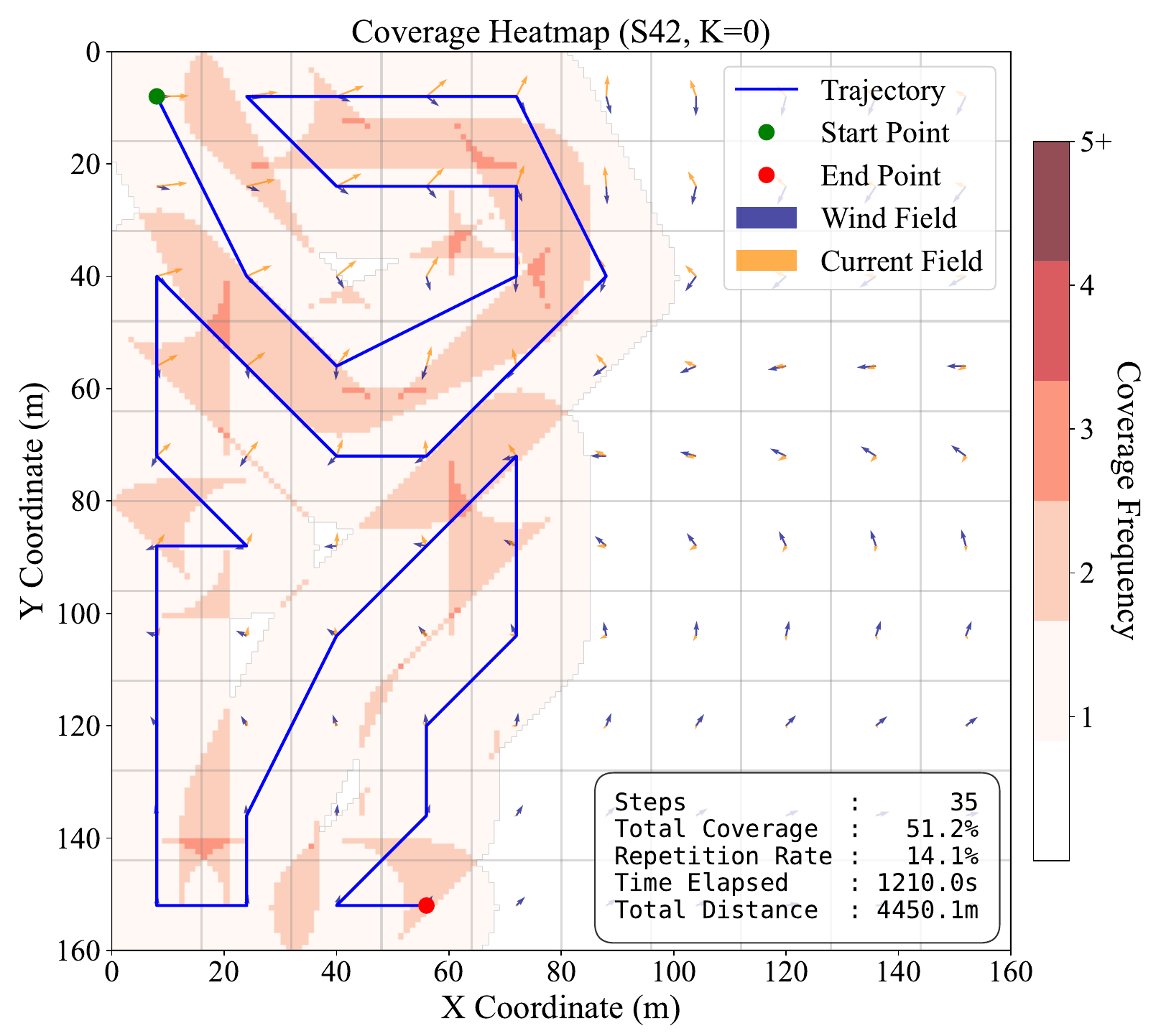}
\hfill
\includegraphics[width=0.31\linewidth]{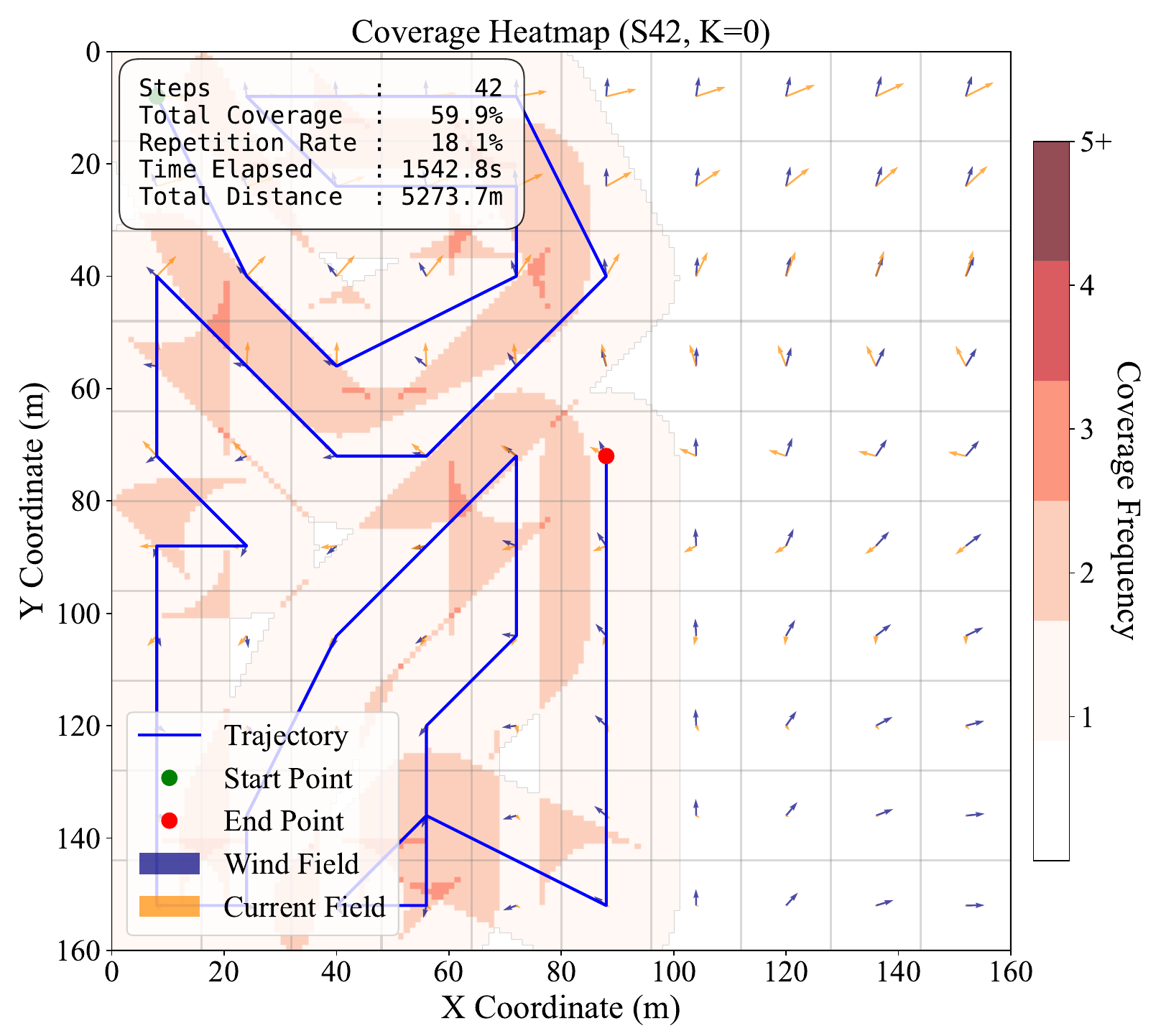}
\hfill
\includegraphics[width=0.31\linewidth]{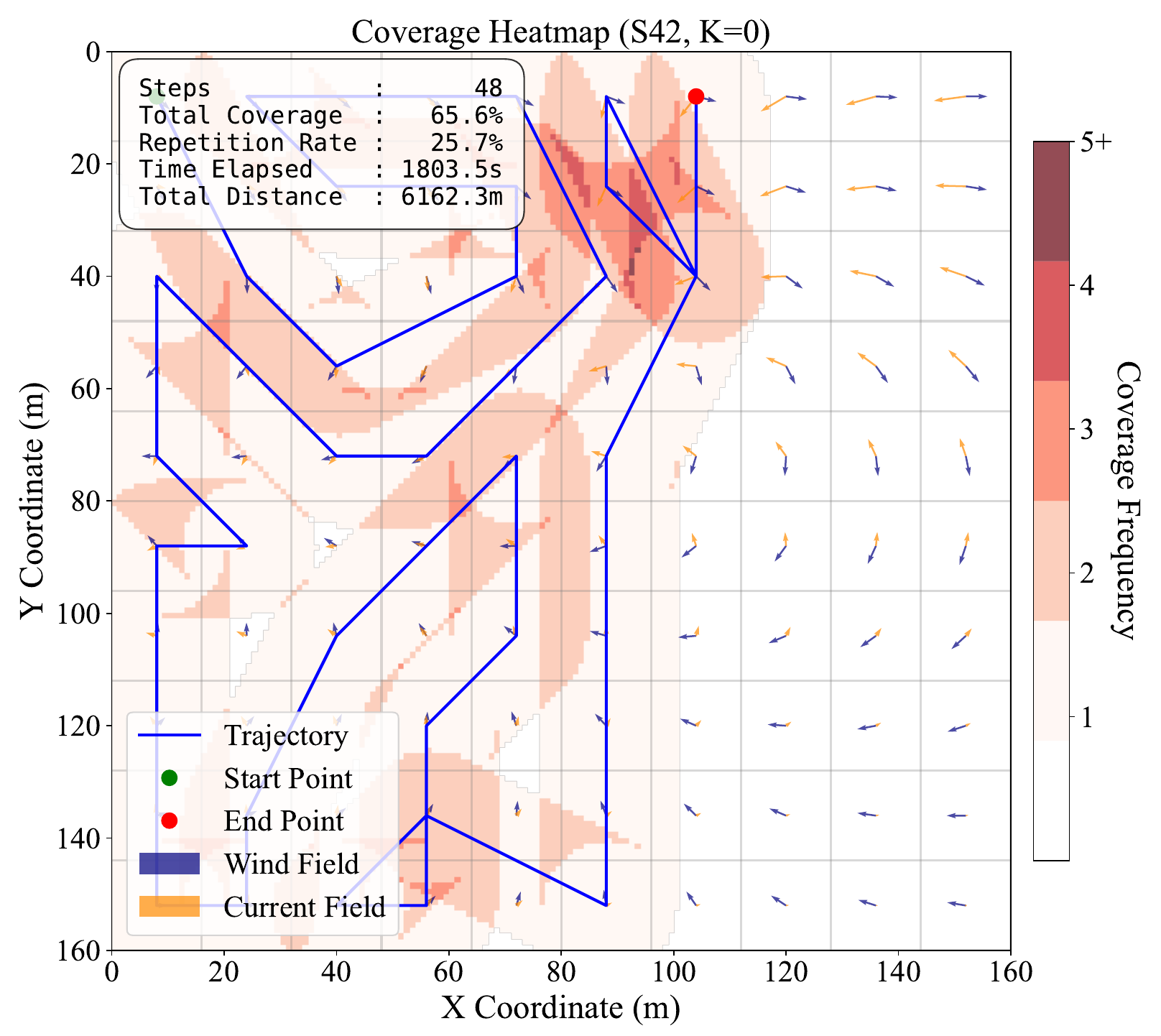}

\vspace{1mm}

\begin{minipage}[t]{0.31\linewidth}
    \centering
    \includegraphics[width=\linewidth]{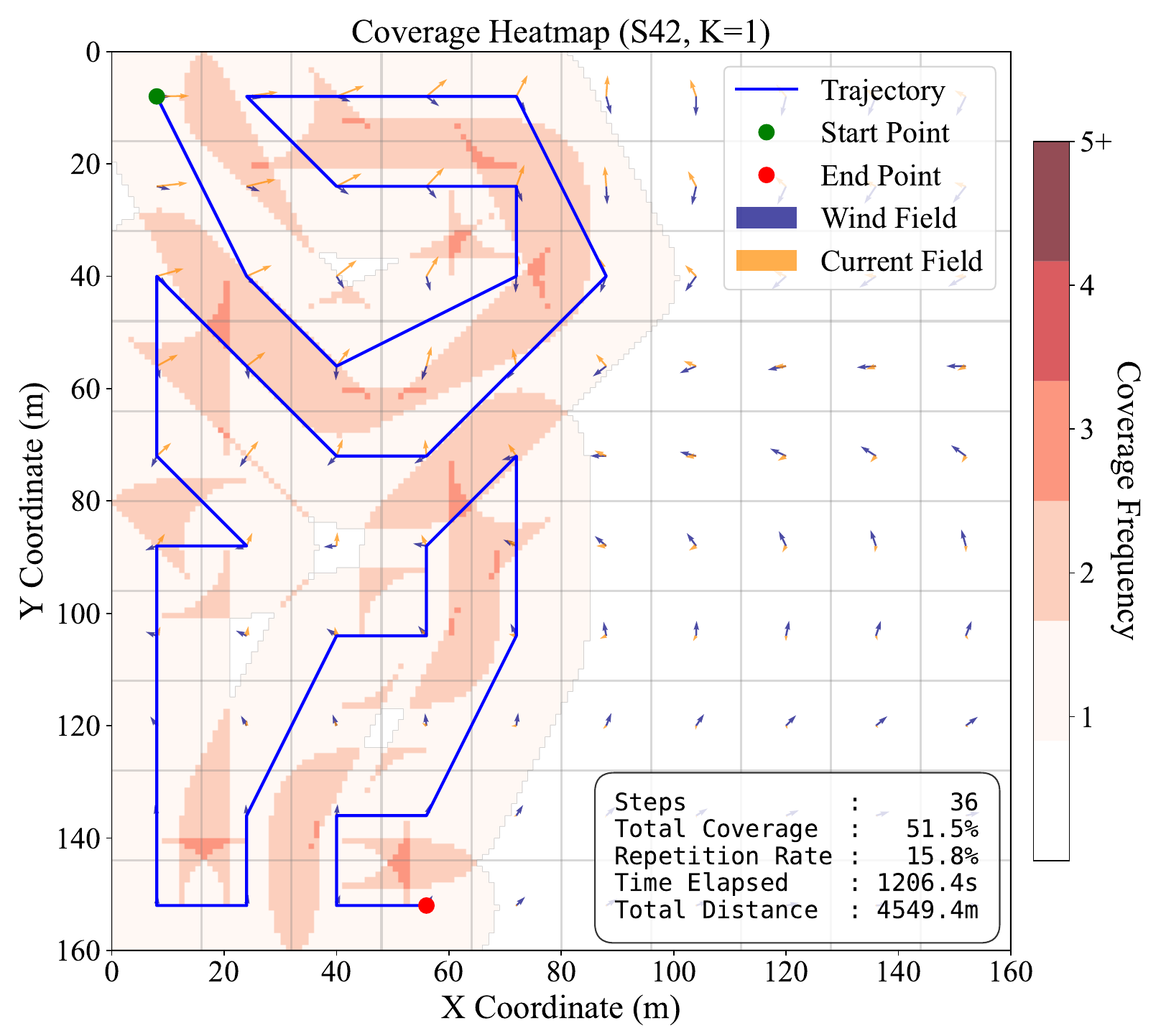}
    
    \vspace{1mm}
    \footnotesize Stage 4
\end{minipage}
\hfill
\begin{minipage}[t]{0.31\linewidth}
    \centering
    \includegraphics[width=\linewidth]{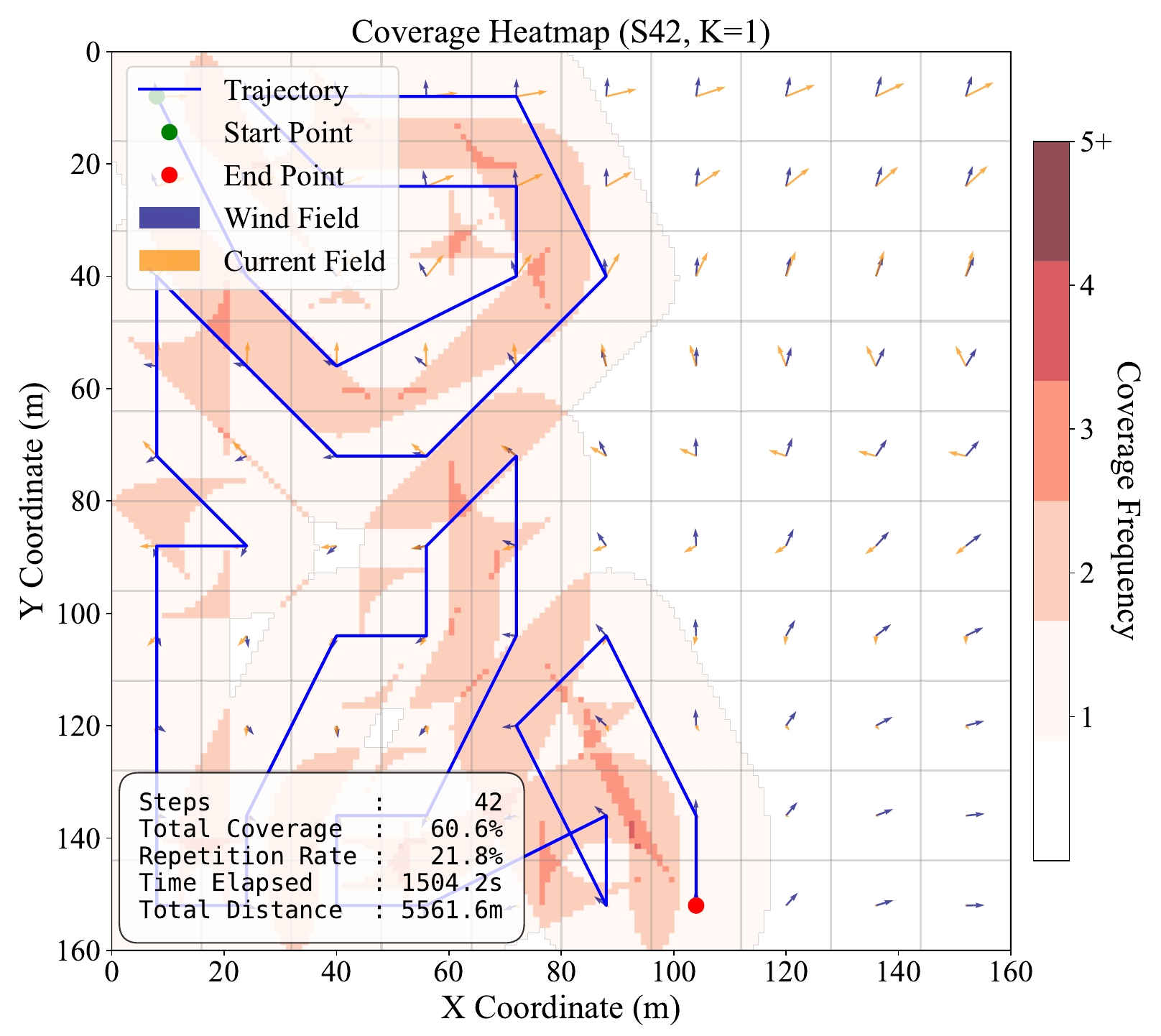}
    
    \vspace{1mm}
    \footnotesize Stage 5
\end{minipage}
\hfill
\begin{minipage}[t]{0.31\linewidth}
    \centering
    \includegraphics[width=\linewidth]{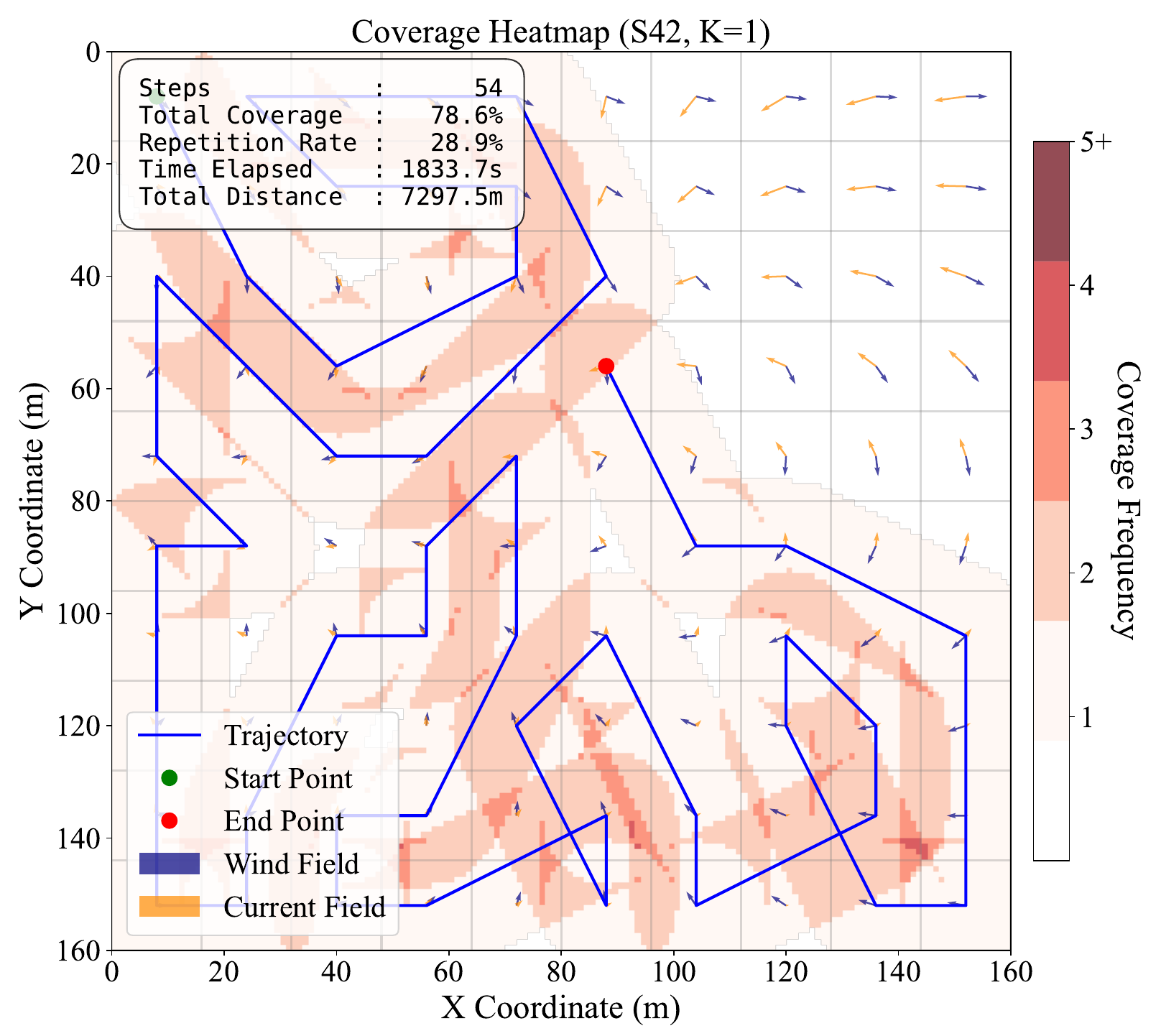}
    
    \vspace{1mm}
    \footnotesize Stage 6
\end{minipage}

\caption{Detailed coverage process of S42 (Stage 4--6). In stages without infeasible regions, the baseline maintains stable and efficient coverage. $K{=}0$ heads upward in Stage~5 but is hindered by adverse currents in Stage~6. In contrast, $K{=}1$ anticipates favorable conditions and achieves more efficient coverage in the lower-right.}

\label{FIG:detail2}
\end{figure*}

\begin{figure*}[!t]
\centering

\includegraphics[width=0.31\linewidth]{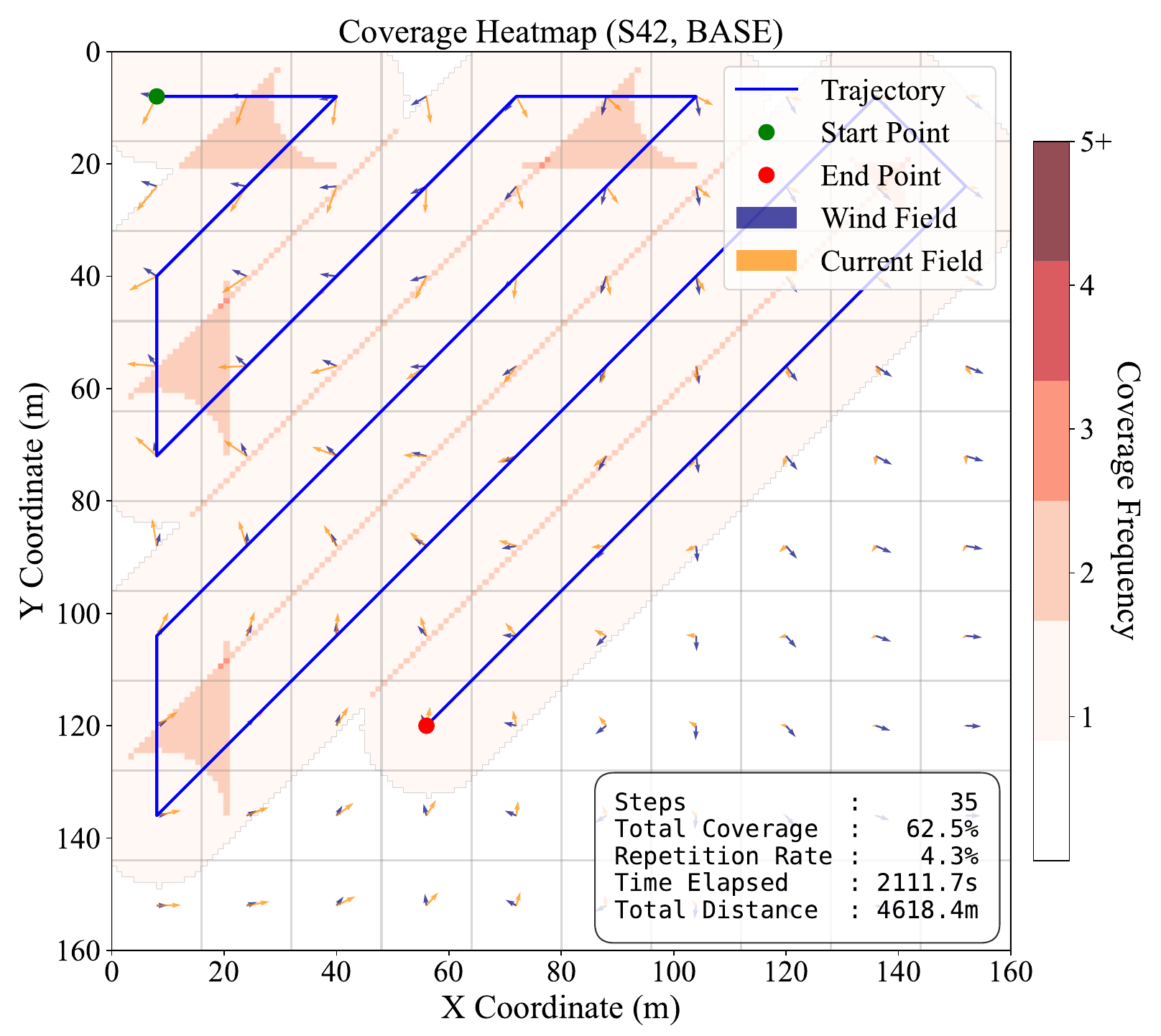}
\hfill
\includegraphics[width=0.31\linewidth]{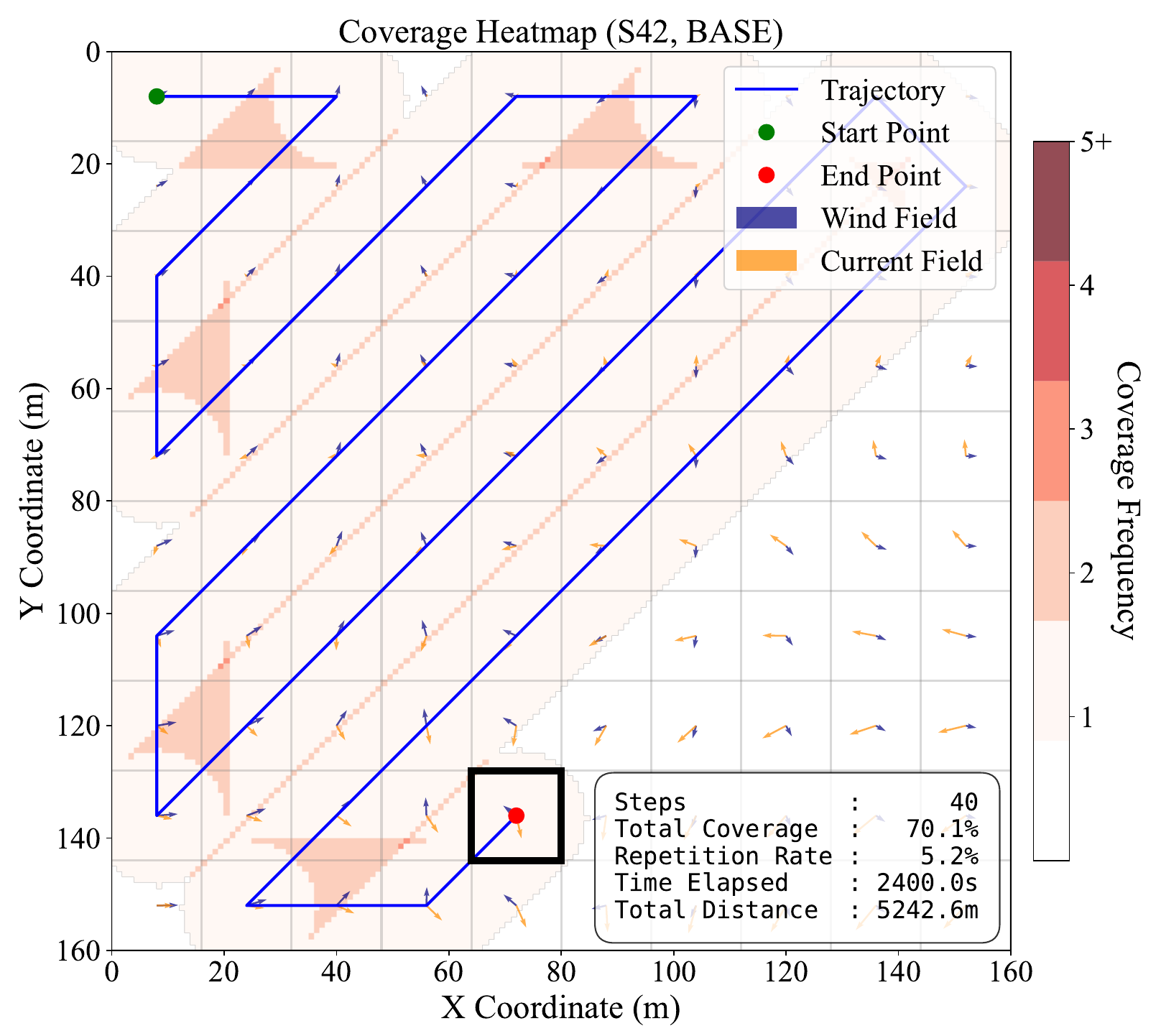}
\hfill
\includegraphics[width=0.31\linewidth]{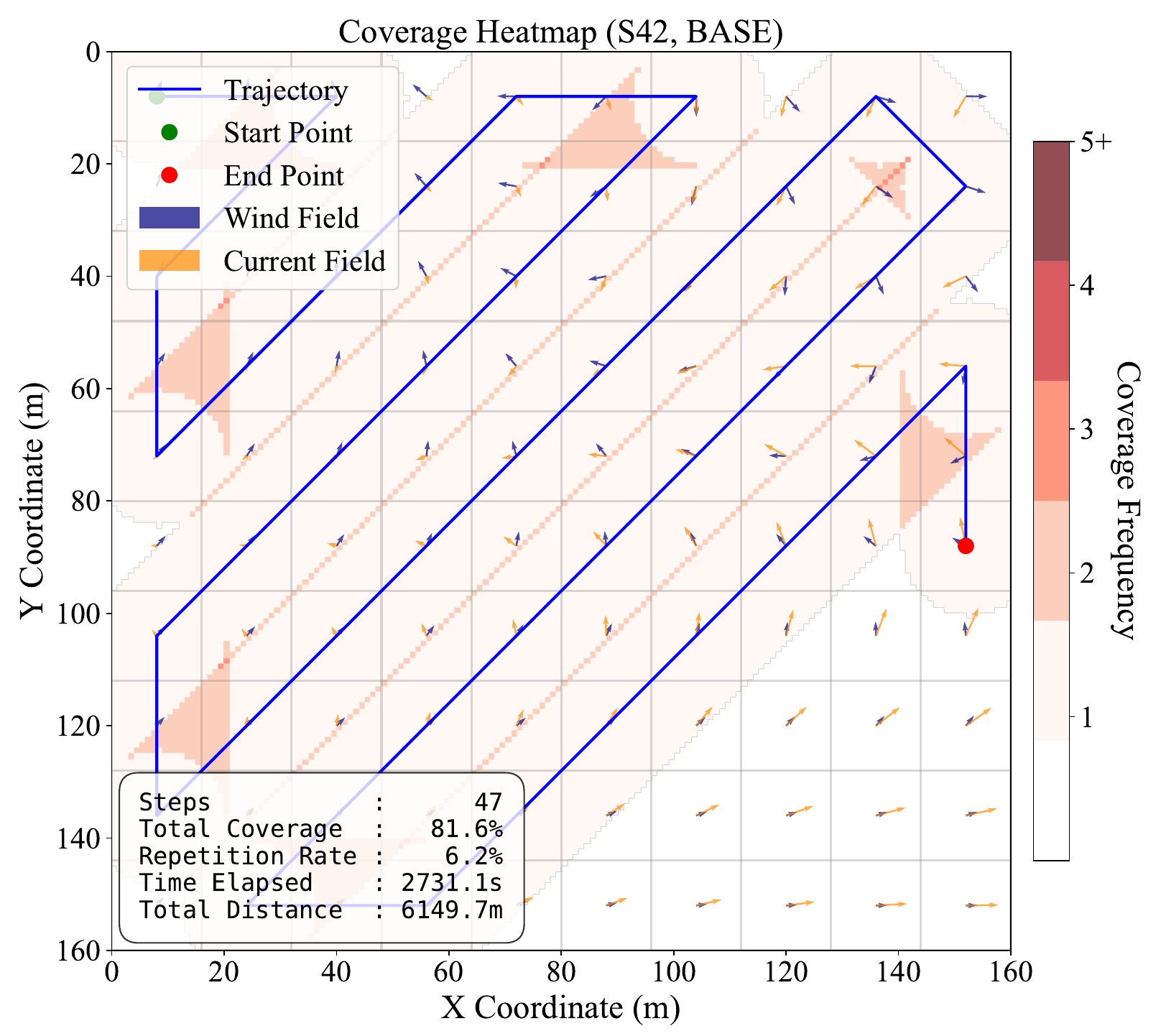}

\vspace{1mm}

\includegraphics[width=0.31\linewidth]{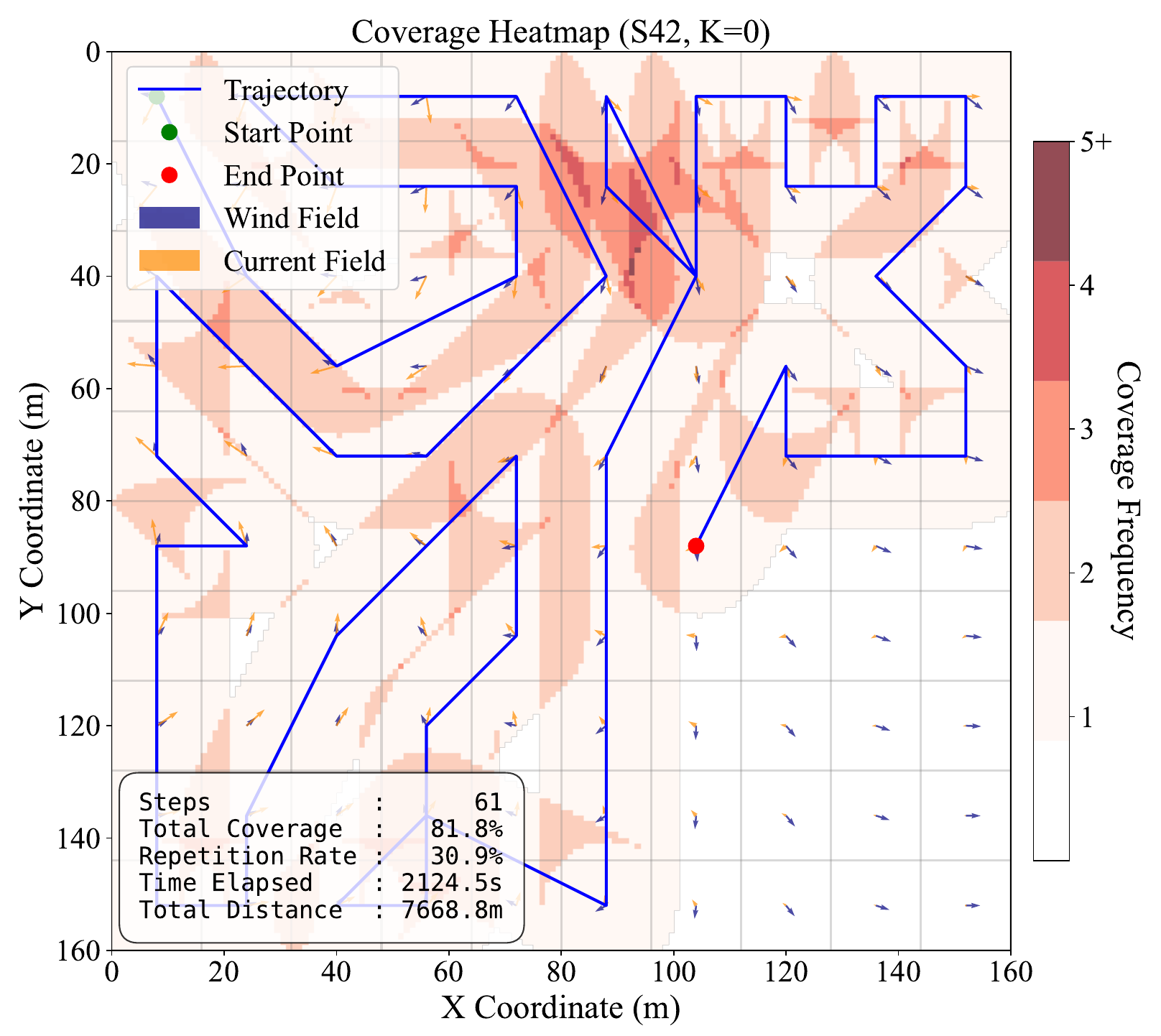}
\hfill
\includegraphics[width=0.31\linewidth]{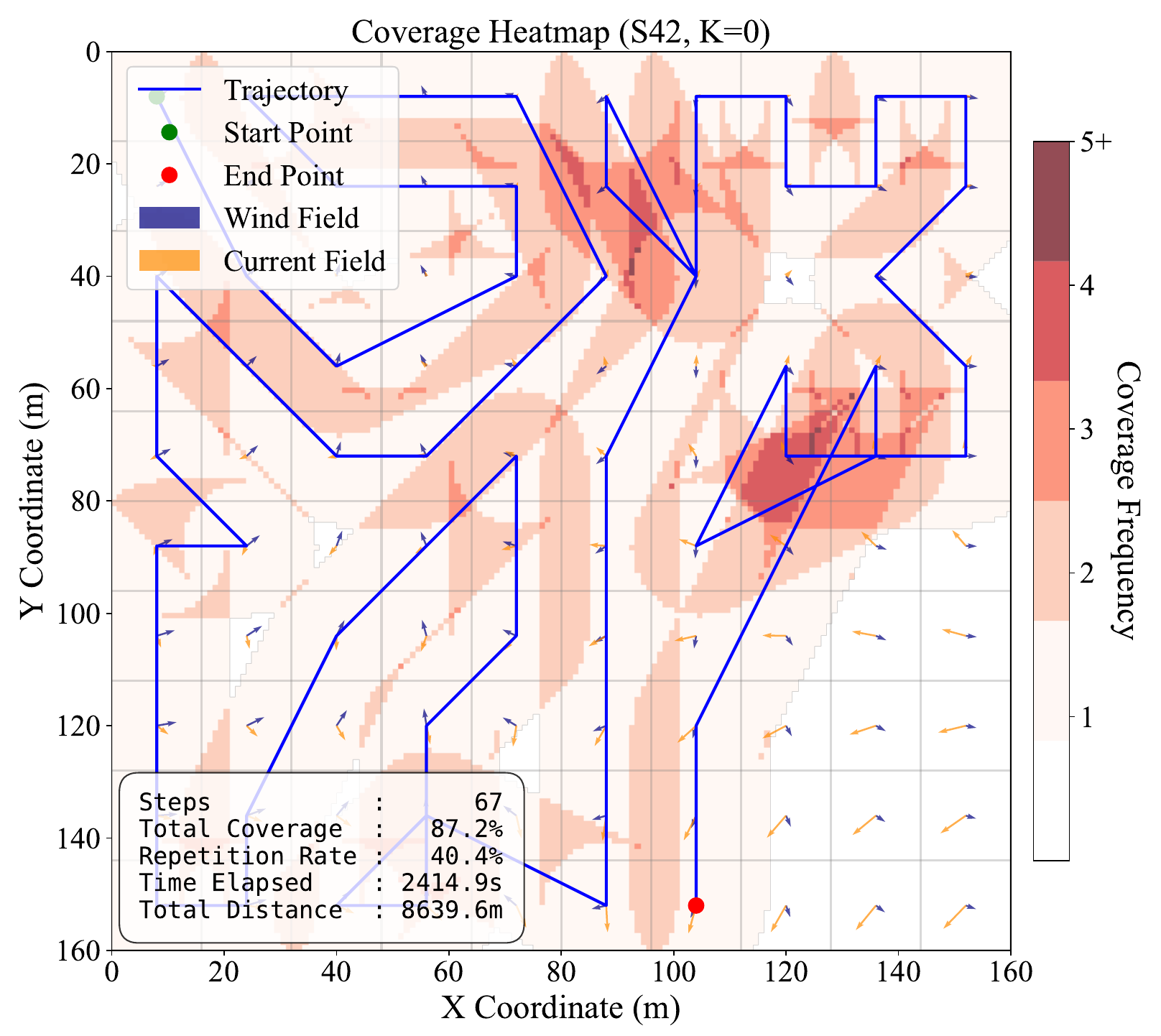}
\hfill
\includegraphics[width=0.31\linewidth]{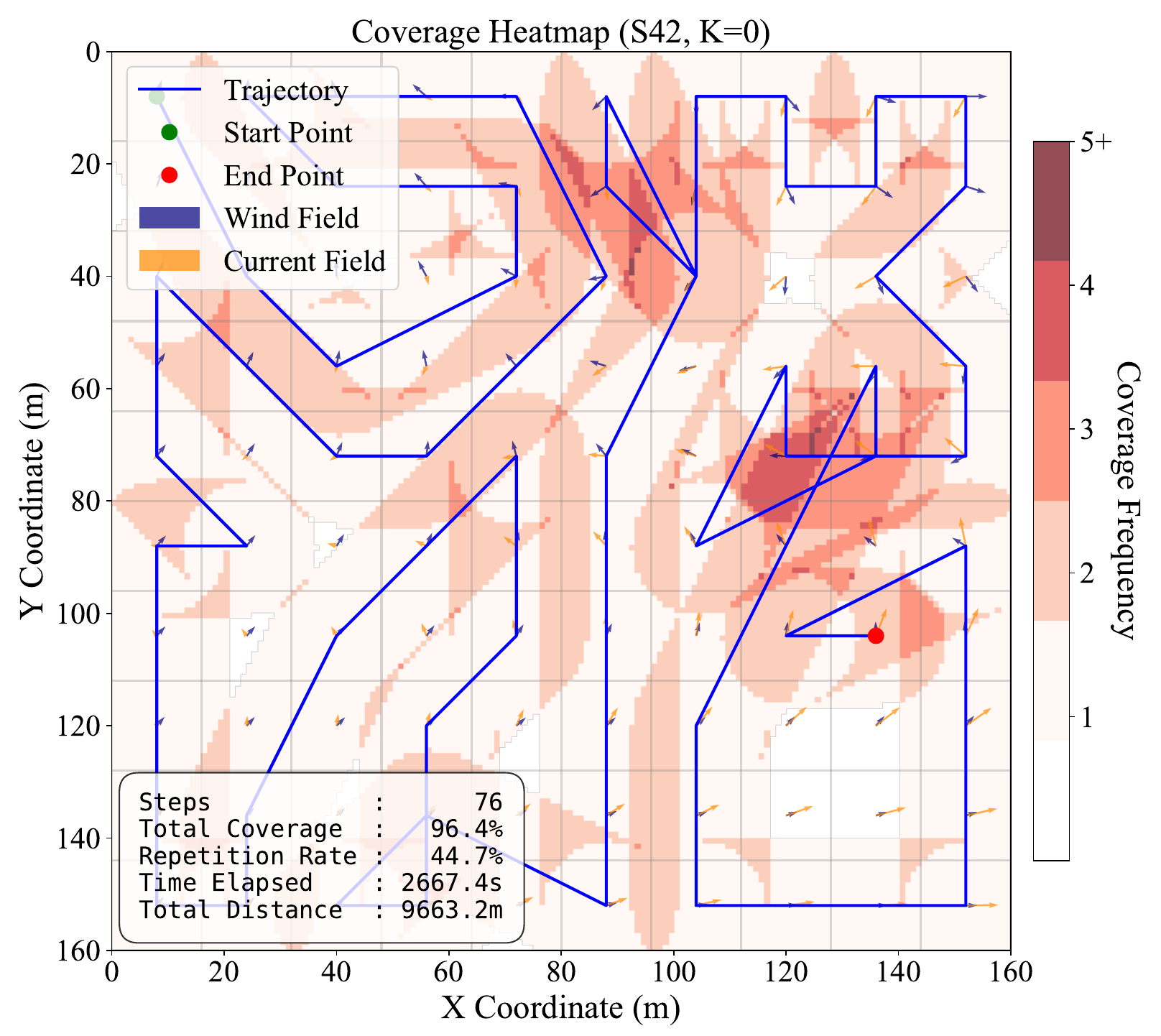}

\vspace{1mm}

\begin{minipage}[t]{0.31\linewidth}
    \centering
    \includegraphics[width=\linewidth]{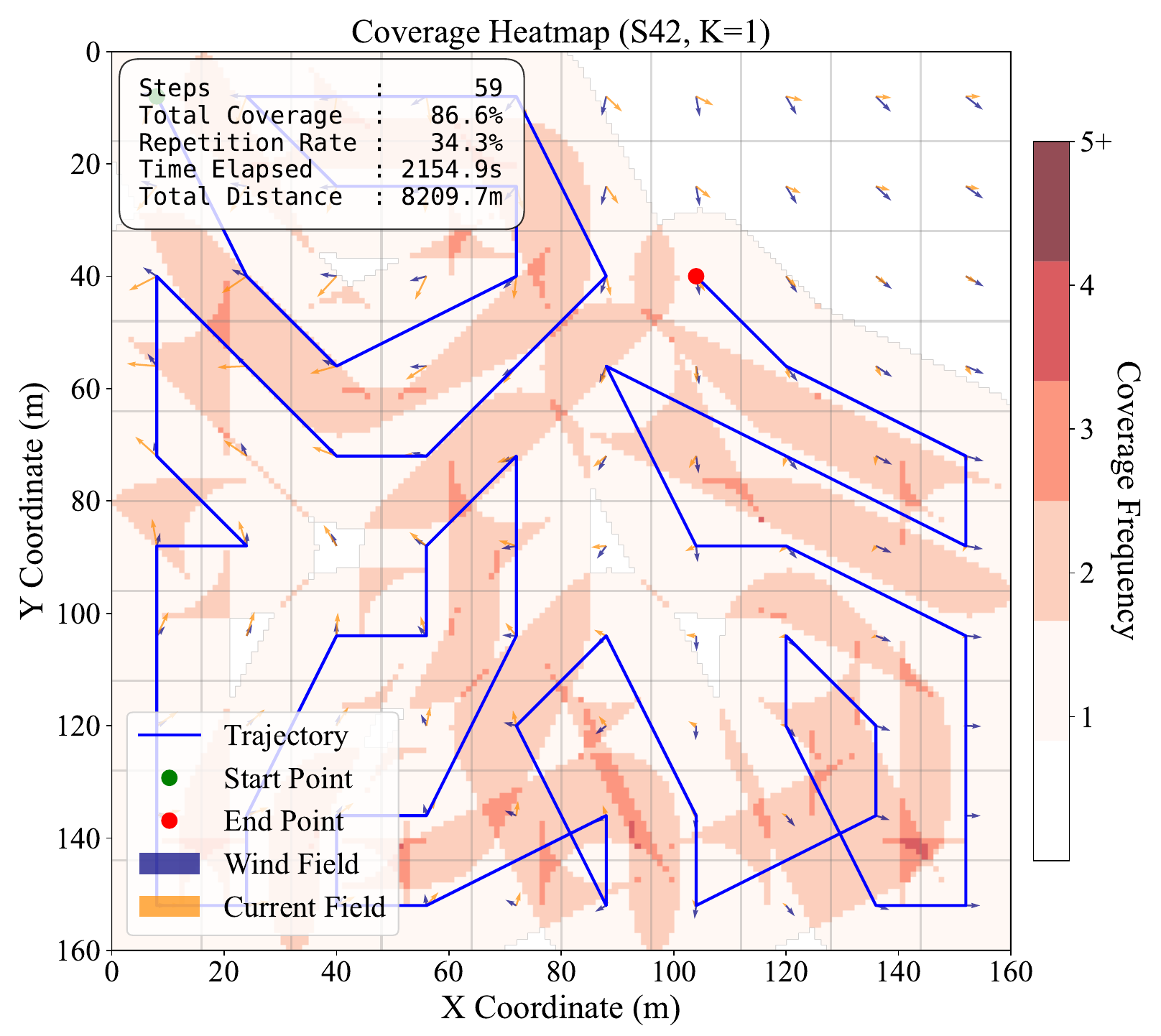}
    
    \vspace{1mm}
    \footnotesize Stage 7
\end{minipage}
\hfill
\begin{minipage}[t]{0.31\linewidth}
    \centering
    \includegraphics[width=\linewidth]{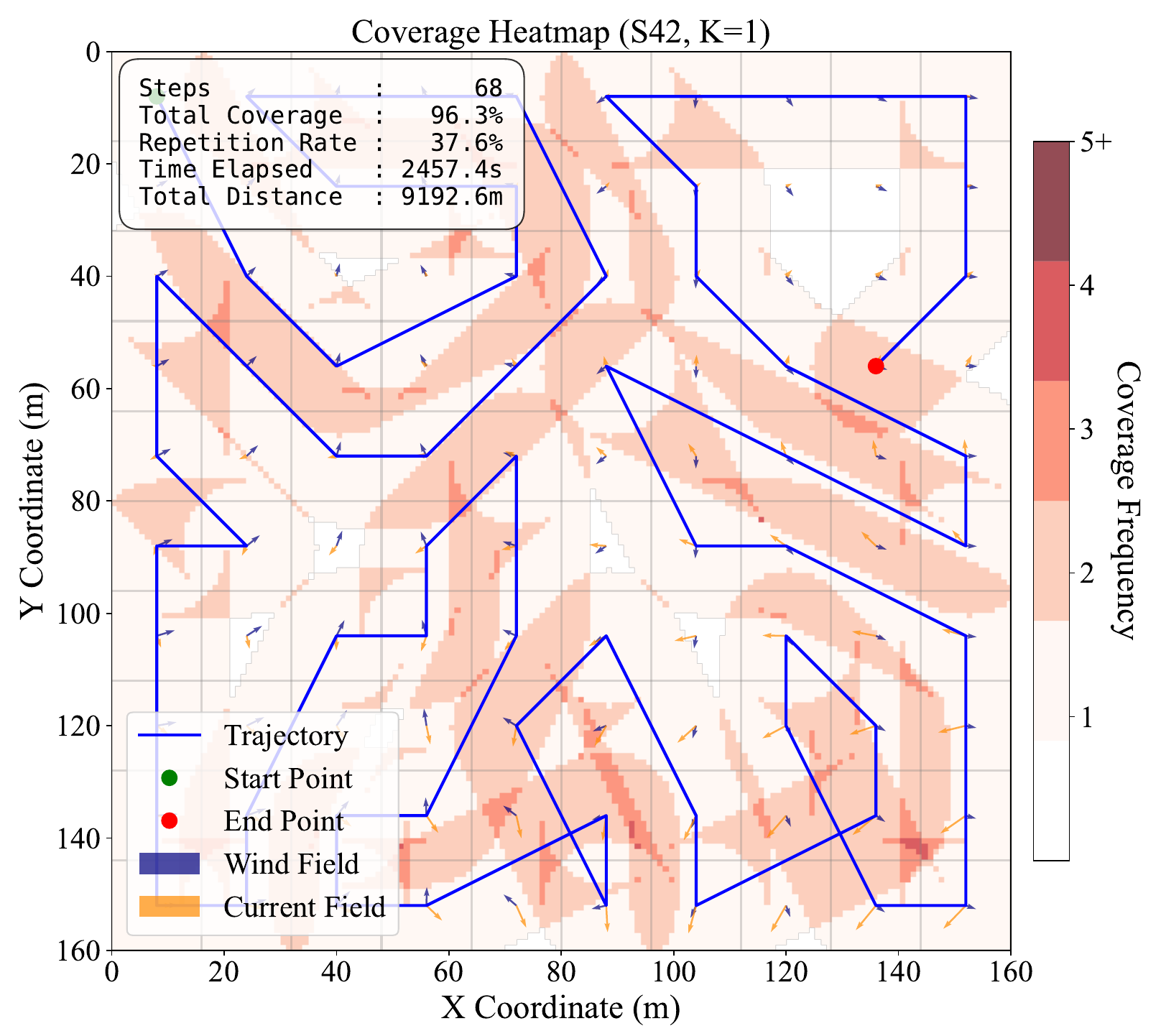}
    
    \vspace{1mm}
    \footnotesize Stage 8
\end{minipage}
\hfill
\begin{minipage}[t]{0.31\linewidth}
    \centering
    \includegraphics[width=\linewidth]{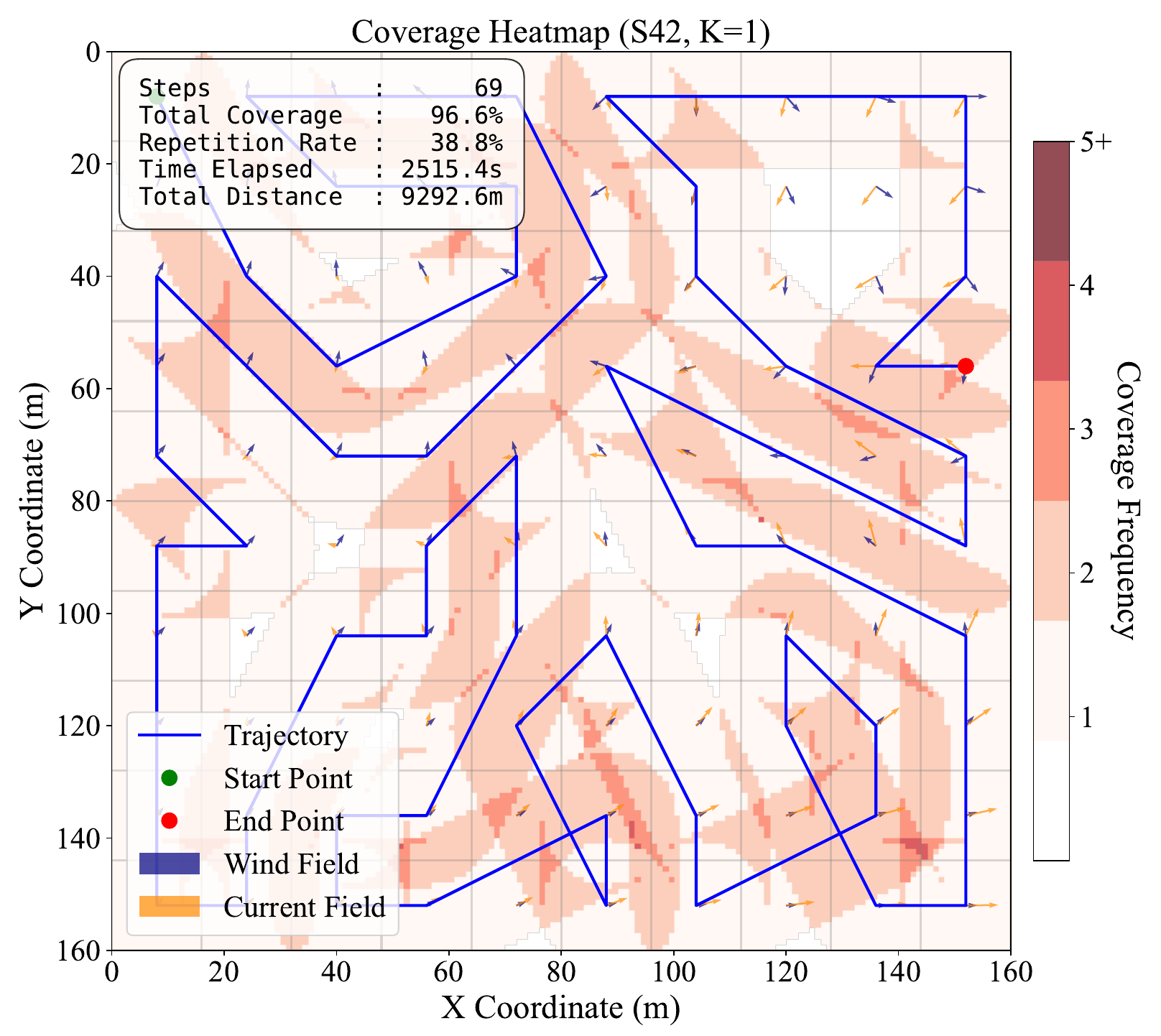}
    
    \vspace{1mm}
    \footnotesize Stage 9
\end{minipage}

\caption{Detailed coverage process of S42 (Stage 7--9). Again, in Stage~8, $K{=}0$ shows limited coverage improvement due to unfavorable conditions, while $K{=}1$ continues to leverage environmental advantages. Compared to the baseline, both proposed methods actively adapt and reach the coverage threshold earlier.}

\label{FIG:detail3}
\end{figure*}

\begin{figure*}[!t]
\centering

\begin{minipage}[t]{0.31\linewidth}
    \centering
    \includegraphics[width=\linewidth]{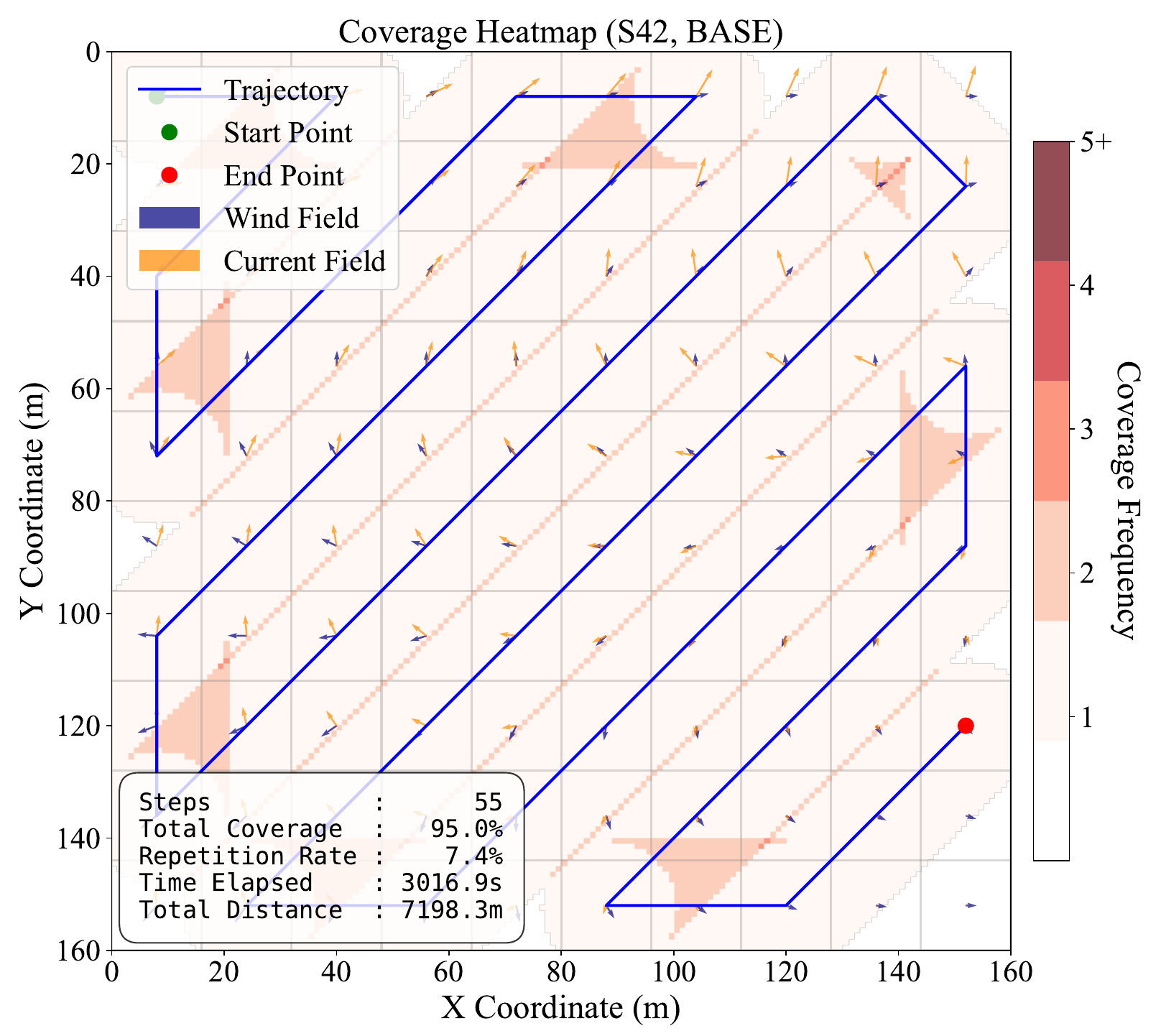}

    \vspace{1mm}
    \footnotesize Stage 10
\end{minipage}
\hfill
\begin{minipage}[t]{0.31\linewidth}
    \centering
    \includegraphics[width=\linewidth]{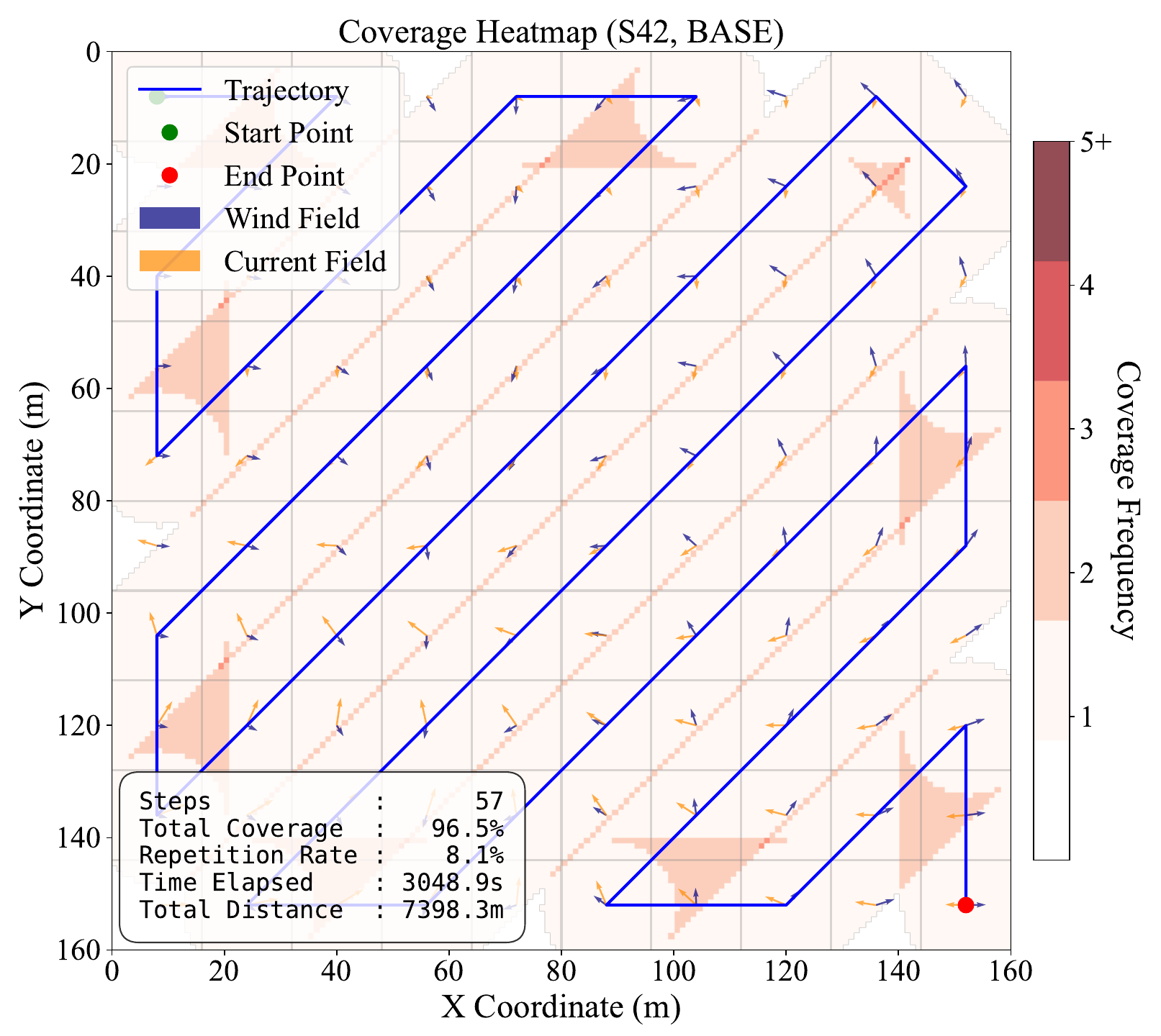}

    \vspace{1mm}
    \footnotesize Stage 11
\end{minipage}
\hfill
\begin{minipage}[t]{0.31\linewidth}
    \centering
    \rule{0pt}{\linewidth}  
\end{minipage}

\caption{Detailed coverage process of S42 (Stage 10--11)}
\label{FIG:detail4}
\end{figure*}

\begin{figure*}[!t]
\centering
\includegraphics[width=\textwidth]{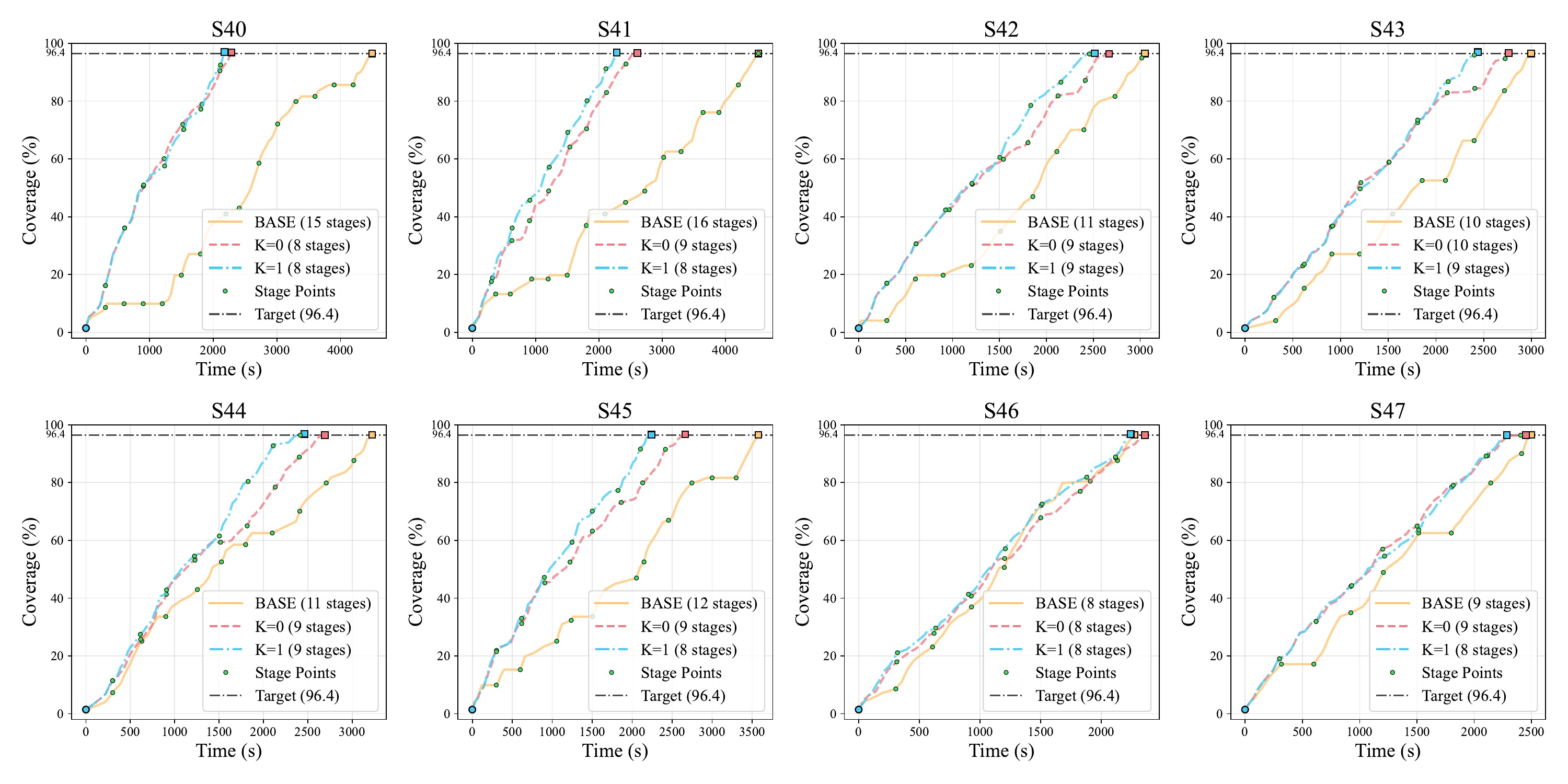}
\caption{Method comparison of time–coverage efficiency (S40--47)}
\label{FIG:result}
\end{figure*}

On average, each planning step with $K=1$ required approximately three times the computation time of $K=0$, reflecting the additional cost of incorporating forecasted intervals. 
Due to space limitations, only the detailed coverage snapshots taken before each environmental update for the baseline (Boustrophedon) and the proposed methods ($K = 0$ and $K = 1$) under S42 are shown in Figs.~\ref{FIG:detail1}--\ref{FIG:detail4} for illustration. The grid boundary that forces the Boustrophedon to pause and wait is highlighted in black. Table~\ref{tab:flow_present_all} summarizes the detailed results. The redundancy metric represents the proportion of the weighted total covered area, consistent with the principle defined in Eq.~\eqref{eq:uniformity}. Figure~\ref{FIG:result} shows the relationship between the coverage ratio and elapsed time for all cases.

\begin{table*}[!t]
\caption{Method comparison under flow-present conditions (S40--47)}
\label{tab:flow_present_all}
\centering
\footnotesize
\renewcommand{\arraystretch}{1.1}
\begin{tabular}{llcccc}
\toprule
\textbf{Case} & \textbf{Method}  & \textbf{Coverage (\%)}$^{\dagger}$ & \textbf{Time Finished (Time await) (s)} & \textbf{Redundancy (\%)} & \textbf{Distance (m)} \\
\midrule

\multirow{3}{*}{S40}
& Base & 96.4 & 4495.9 (1791.8) & 8.1 & 7398.3 \\
& Proposed ($K{=}0$) & 96.9 & 2286.9 & 33.6 & 9004.6 \\
& Proposed ($K{=}1$) & 96.6 & 2177.8 & 27.1 & 8581.7 \\
\midrule

\multirow{3}{*}{S41}
& Base & 96.4 & 4523.7 (1483.7) & 8.1 & 7398.3 \\
& Proposed ($K{=}0$) & 96.7& 2609.9 & 40.8 & 9379.9 \\
& Proposed ($K{=}1$) & 96.9 & 2284.0 & 41.6 & 9462.5 \\

\midrule
\multirow{3}{*}{S42}
& Base & 96.4 & 3048.9 (760.1)  & 8.1 & 7398.3 \\
& Proposed ($K{=}0$) & 96.4 & 2667.4 & 44.7 & 9663.2 \\
& Proposed ($K{=}1$) & 96.6 & 2515.4 & 38.8 & 9292.6 \\
\midrule
\multirow{3}{*}{S43}
& Base & 96.4 & 2997.8 (655.3) & 8.1 & 7398.3 \\
& Proposed ($K{=}0$) & 96.6 & 2761.5 & 48.4 & 9898.4 \\
& Proposed ($K{=}1$) & 97.0 & 2438.4 & 40.7 & 9457.6 \\
\midrule

\multirow{3}{*}{S44}
& Base & 96.4 & 3224.7 (457.0) & 8.1 & 7398.3 \\
& Proposed ($K{=}0$) & 96.4 & 2691.5 & 43.6 & 9550.5 \\
& Proposed ($K{=}1$) & 96.8 & 2463.5 & 47.9 & 9928.9 \\
\midrule
\multirow{3}{*}{S45}
& Base & 96.4 & 3578.5 (990.2) & 8.1 & 7398.3 \\
& Proposed ($K{=}0$) & 96.7 & 2662.9 & 40.8 & 9474.8\\
& Proposed ($K{=}1$) & 96.6 & 2240.4 & 38.5 & 9251.8 \\
\midrule
\multirow{3}{*}{S46}
& Base & 96.4 & 2273.4 (39.7) & 8.1 & 7398.3 \\
& Proposed ($K{=}0$) & 96.4 & 2357.9 & 40.4 & 9357.6 \\
& Proposed ($K{=}1$) & 96.8 & 2241.2 & 31.4 & 8828.9 \\
\midrule
\multirow{3}{*}{S47}
& Base & 96.4 & 2496.9 (568.7) & 8.1 & 7398.3 \\
& Proposed ($K{=}0$) & 96.4 & 2451.4 & 42.9 & 9588.1 \\
& Proposed ($K{=}1$) & 96.4  & 2284.6 & 36.4 & 9117.3 \\

\midrule

\multirow{3}{*}{Avg $\pm$ Std$^{*}$}
& Base & \textemdash & 3330.0 {\scriptsize$\pm$833.0} (843.3 {\scriptsize$\pm$566.2}) & 8.1 {\scriptsize$\pm$0} & 7398.3 {\scriptsize$\pm$0} \\
& Proposed ($K{=}0$) & \textemdash & 2561.2 {\scriptsize$\pm$173.1} & 41.9 {\scriptsize$\pm$4.3} & 9489.6 {\scriptsize$\pm$260.6} \\
& Proposed ($K{=}1$) &  \textemdash & 2330.7 {\scriptsize$\pm$123.7} & 37.8 {\scriptsize$\pm$6.4} & 9240.2 {\scriptsize$\pm$412.6} \\
\bottomrule
\end{tabular}

\vspace{0.5em}
\footnotesize
\textit{Note:} $^{\dagger}$ Coverage is measured up to the point where boustrophedon coverage is reached or exceeded in each case.\\
$^{*}$ Reported statistics reflect scenario-level trends and may not carry physical significance due to environmental variability.

\end{table*}

From Table~\ref{tab:flow_present_all}, it can be seen that the proposed method successfully generates valid coverage paths across all flow-present scenarios (S40--S47), regardless of whether $K{=}0$ or $K{=}1$ is used. Compared to the baseline boustrophedon path, it achieves a substantial improvement in the most critical metric—coverage time. On average, the proposed method with $K{=}0$ reduces the coverage time from 3330.0~s to 2561.2~s, while the $K{=}1$ configuration further lowers it to 2330.7~s, demonstrating the benefit of lookahead-based planning.

A special case occurs in S46, where the randomly generated environment is particularly favorable for the boustrophedon strategy, resulting in a total waiting time of only 39.7~s. Under such conditions, the gap-free and non-overlapping characteristics of the baseline path are fully exploited. Although the individual maneuvers are not globally optimized, their cumulative effective coverage remains consistently high, as also illustrated in Fig.~\ref{FIG:result}. Nevertheless, even in this favorable setting, the proposed method still returns feasible paths with competitive performance.

The redundancy metric has a dual interpretation. While increased redundancy may suggest inefficient path use, it can also enhance robustness and data reliability in practical ocean observation tasks. The proposed method incurs higher redundancy (e.g., 41.9\% for $K{=}0$ and 37.8\% for $K{=}1$ versus 8.1\% for the baseline), but this is accompanied by significantly shorter mission times and more adaptive global coverage.

Although the proposed trajectories are generally longer than the baseline (e.g., 9240.2~m for $K{=}1$ versus 7398.3~m), such increased sailing distances are not necessarily disadvantageous for autonomous sailboats. Since propulsion relies on wind—a renewable energy source—longer paths may actually reflect better routing decisions that maintain higher average speeds by leveraging favorable environmental flows.

The statistical results also indicate that the $K{=}1$ configuration consistently outperforms $K{=}0$ in all scenarios. Its single-step lookahead capability enables better foresight and adaptive planning, which is especially evident in the detailed analysis of Case S42 (Figs.~\ref{FIG:detail1}--\ref{FIG:detail4}). During the initial stages, both variants produce similar paths due to shared heuristics and rollout logic. However, from Stage~5 onward, $K{=}0$, guided only by short-term gains, moves toward the upper-right region, eventually encountering adverse currents that limit its coverage gain (+5.7\% in Stage~6). In contrast, $K{=}1$ steers toward the lower-right, anticipating a favorable current that yields a significantly higher coverage gain (+18.0\%) in the same stage. This trend continues in later stages, where $K{=}1$ maintains superior performance by consistently avoiding strong counter-flows.

In summary, the proposed spatiotemporal planning framework demonstrates strong adaptability and efficiency under dynamic marine conditions. It reliably produces high-quality coverage solutions across all test cases, and the temporal lookahead strategy (\( K{=}1 \)) further enhances performance by enabling the planner to anticipate environmental transitions and make globally beneficial decisions.

\section{Conclusion}

This paper addresses the area coverage problem for autonomous sailboats operating under time-varying wind and current fields. A practical problem formulation is presented to capture real-world maneuvering and sensing constraints. Building on this formulation, a spatiotemporal coverage path planning framework is developed by integrating morphological constraints in the spatial domain with look-ahead planning in the temporal domain. Comparative experiments with a baseline boustrophedon path in randomly generated ocean scenarios demonstrate that the proposed method yields efficient and feasible coverage paths, even when partial inaccessibility may occur due to adverse flow conditions or directional sailing limitations. The resulting framework offers a practical foundation for adaptive, long-duration ocean observation tasks using autonomous sailboats, and provides a basis for future research on cooperative multi-sailboat coverage. As a next step, we plan to extend this work toward multi-sailboat collaboration in inhomogeneous and time-varying ocean environments, aiming to further enhance efficiency and resilience in large-scale missions.

\bibliographystyle{IEEEtran}
\bibliography{IEEEabrv,ref}

\vfill

\end{document}